\documentclass[journal]{IEEEtran}
\usepackage[T1]{fontenc}
\usepackage{amsmath,amssymb,amsfonts}
\usepackage{algorithmic}
\usepackage{array}
\usepackage[caption=false,font=normalsize,labelfont=sf,textfont=sf]{subfig}
\usepackage{stfloats}
\usepackage{url}
\usepackage{graphicx}
\usepackage{xcolor}
\usepackage{multirow}
\usepackage{enumitem}
\usepackage{arydshln}
\usepackage{booktabs}
\usepackage{pifont}
\usepackage{booktabs}
\usepackage{pifont}
\usepackage{makecell}
\newcommand{\xmark}{\ding{55}}
\usepackage{makecell}
\usepackage{mathrsfs}
\usepackage{balance}
\usepackage[numbers,sort&compress]{natbib}
\usepackage{hyperref}

\usepackage{paracol}
\usepackage{capt-of}

\definecolor{mypink}{RGB}{255, 105, 180}  

\hypersetup{
    colorlinks=true,  
    urlcolor=mypink,  
}

\def\BibTeX{{\rm B\kern-.05em{\sc i\kern-.025em b}\kern-.08em
    T\kern-.1667em\lower.7ex\hbox{E}\kern-.125emX}}
\newcommand{\cmark}{\ding{51}}

\begin{document} 

\title{A Unified Benchmark and Modality-Adaptive Network for 
Day-Night Drone-View Geo-Localization Tasks}

\author{
Songtianhao Xu, Zhongwei Chen,
Hang Wang, Weifeng Wang, Zhao-Xu Yang, \textit{Member, IEEE}\\

\thanks{This paper is submitted for review on \today. This work was supported in part by the Key Research and Development Program of Shaanxi, PR China (No. 2025CY-JJQ-189), Key Scientific and Technological Projects of Henan Province (No. 252102241050), the National Natural Science Foundation of China (No. 61976172 and No. 12002254). Songtianhao Xu and Zhongwei Chen contributed equally to this work. (Corresponding authors: Zhao-Xu Yang and Weifeng Wang)
}

\thanks{
Zhongwei Chen and Zhao-Xu Yang are with the State Key Laboratory for Strength and Vibration of Mechanical Structures, Shaanxi Key Laboratory of Environment and Control for Flight Vehicle, School of Aerospace Engineering, Xi'an Jiaotong University, Xi'an 710049, PR China (e-mail: ISChenawei@stu.xjtu.edu.cn; yangzhx@xjtu.edu.cn).
}
\thanks{
Hang Wang is with the State Key Laboratory of Intelligent Agricultural Power Equipment, College of Vehicle and Traffic Engineering, Henan University of Science and Technology, Luoyang 471003, PR China (e-mail: wanghang@haust.edu.cn).
}
\thanks{
Songtianhao Xu and Weifeng Wang are with the Xi'an Institute of Optics and Precision Mechanics, Chinese Academy of Sciences, Xi'an 710119, PR China (e-mail: xusongtianhao26@mails.ucas.ac.cn; wwf7911@opt.ac.cn).
}

}

\maketitle

\begin{abstract}
Most existing drone-view geo-localization (DVGL) benchmarks collect aerial images under limited illumination variations and lack geographically aligned multimodal data, which encompass visible aerial, infrared aerial, and satellite reference images captured at identical geographic locations. To evaluate the cross-modality generalizability of DVGL methods for day-night tasks, researchers commonly train models on a visible benchmark and test them on an independent infrared benchmark. This protocol relies on cross-dataset transfer, which makes it difficult to systematically evaluate DVGL methods during the day and at night within a unified benchmark. To address this limitation, a real-world cross-modality DVGL benchmark termed IRVSGeo is introduced, which involves visible aerial, infrared aerial, and satellite reference images. IRVSGeo contains 26,460 images collected from 8,820 geographic locations across four representative scene categories, which include cropland, forest, urban, and coastal areas. Each location provides one visible aerial image captured under diverse illumination conditions, one infrared aerial image, and one corresponding satellite reference image. Furthermore, the modality-adaptive state-space transport relation network (MASTR-Net) is proposed, which not only supports conventional unimodal training and inference, but also enables joint training of multimodal query or gallery images for modality-adaptive unimodal inference, namely using visible aerial images for daytime geo-localization tasks, and infrared aerial images for nighttime geo-localization tasks when visible observations are unavailable. Extensive experiments demonstrate that MASTR-Net outperforms state-of-the-art methods on IRVSGeo across various tasks and achieves competitive results on two infrared benchmarks, IR-VL328 and CVGL-RGBT. Code: \textcolor{magenta}{\url{https://github.com/SongtianhaoXu/MASTR-Net}}.

\end{abstract}


\section{Introduction}

\begin{figure}[ht]
    \centering
    \includegraphics[width=\columnwidth]{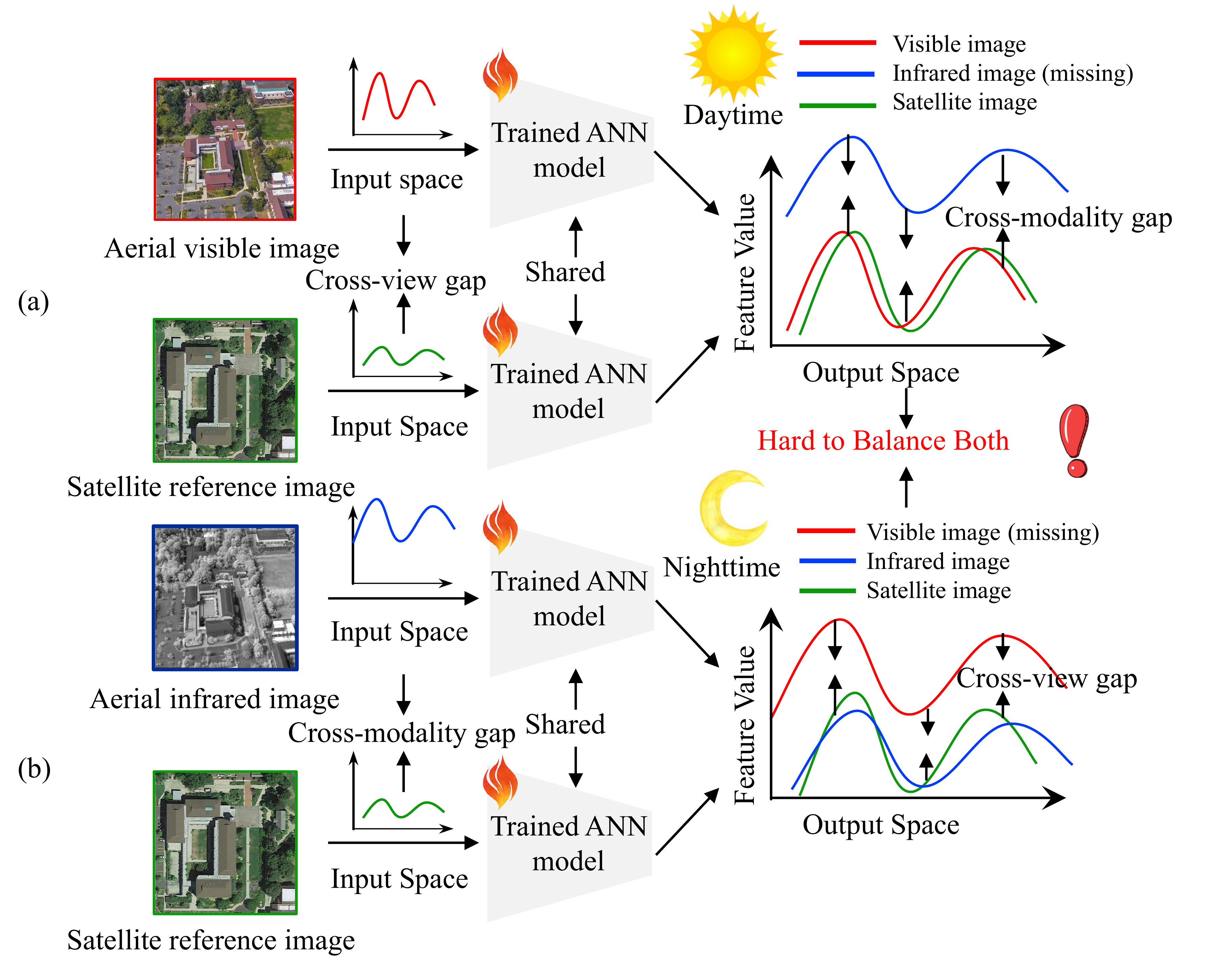}
    \caption{\textbf{Motivation for day-night DVGL tasks.}
   (a) For daytime tasks, current methods typically use a shared trained artificial neural network (ANN) model to match visible aerial images with satellite reference images and bridge the cross-view gap. (b) For nighttime tasks, a shared model is similarly used to match infrared aerial images with satellite reference images and mitigate the cross-modality gap. However, the models for these two tasks are trained independently, and their matching capabilities do not readily transfer between the two tasks. There remains a lack of a unified model with cross-modality generalization that can effectively reconcile daytime and nighttime performance. Cross-modality generalization between visible and infrared modalities offers a viable solution to address challenging illumination conditions, as it mitigates the limitations of visible imagery under low-illumination conditions and accommodates the inherent artifacts and radiometric properties of infrared imagery. Unfortunately, progress in this field is constrained by the absence of a unified benchmark with geographically aligned visible aerial, infrared aerial, and satellite reference images, which prevents systematic and consistent evaluation of daytime and nighttime geo-localization.}
\label{fig1}
\end{figure}

\IEEEPARstart{D}{rone-view} geo-localization (DVGL) aims to infer the geographic location of a target region through a retrieval process where a drone-view query image is matched with geo-tagged satellite reference images \cite{chen2024multi,xia2024enhancing,zheng2020university}. As one of the vision-based geo-localization techniques independent of GNSS signals, DVGL is particularly valuable for drone flight in GNSS-denied scenarios {\cite{dai2023vision,durgam2024cross,chen2026geomfd}}. In recent years, substantial progress has been reported on several visible-spectrum DVGL benchmarks \cite{shen2023mccg,deuser2023sample4geo,zhu2023sues}. However, current methods are mostly restricted to daytime tasks that use only visible images \cite{11622533,zhao2024transfg,chen2024sdpl}. As illustrated in Fig. \ref{fig1}(a), most public benchmarks contain only visible aerial images captured under favorable illumination conditions and their corresponding satellite reference images, while infrared data acquired at night or under low-illumination conditions are rarely included \cite{zheng2020university,zhu2023sues,dai2023vision}. Consequently, current methods mainly rely on visible cues, such as color, texture, edges, and local appearance, to perform cross-view retrieval that matches visible aerial images with satellite reference images of the same locations {\cite{wang2021each,lin2022joint}}. Although these cues are robust under favorable illumination, they degrade severely at night or under low-illumination conditions. Given that practical drone platforms switch between visible and infrared sensors to support persistent day-night geo-localization, such illumination-sensitive cues inevitably lead to less discriminative features and unreliable cross-view matching.

Visible and infrared aerial images provide complementary observations for day-night geo-localization tasks. Infrared imagery remains informative when low illumination degrades visible imagery, but the appearance discrepancies between the two modalities require a model capable of processing both within a unified framework. To assess the transfer of DVGL models from visible to infrared imagery under challenging illumination conditions, some studies adopted a protocol, in which models trained on a visible benchmark were directly evaluated on an independent infrared benchmark \cite{liu2025object}. However, this protocol essentially evaluates cross-dataset transfer and is also confounded by differences in scene content, acquisition conditions, and data distributions. It therefore cannot isolate the effects of modality and illumination variations on geo-localization performance. Meanwhile, as shown in Fig.~\ref{fig1}(b), existing infrared DVGL benchmarks provide infrared aerial images for model training and testing, but usually lack geographically aligned visible aerial images \cite{zhou2025cdm,liu2025object}. As a direct consequence, the absence of paired visible imagery at the same geographic locations makes it impossible to conduct fair head-to-head comparison of visible and infrared geo-localization performance.

Moreover, DVGL under varying illumination conditions must simultaneously address the cross-view gap and the cross-modality gap {\cite{zhang2024geodtr+,cui2021cross}}. Specifically, the DVGL model must resolve the cross-view structural discrepancies between images captured by drone and satellite, and bridge the cross-modality appearance discrepancies between visible and infrared aerial imagery. It is essential to establish a geographically aligned multimodal benchmark and develop a model capable of jointly addressing these coupled gaps under varying illumination conditions.

\begin{figure}[t]
    \centering
    \includegraphics[width=\columnwidth]{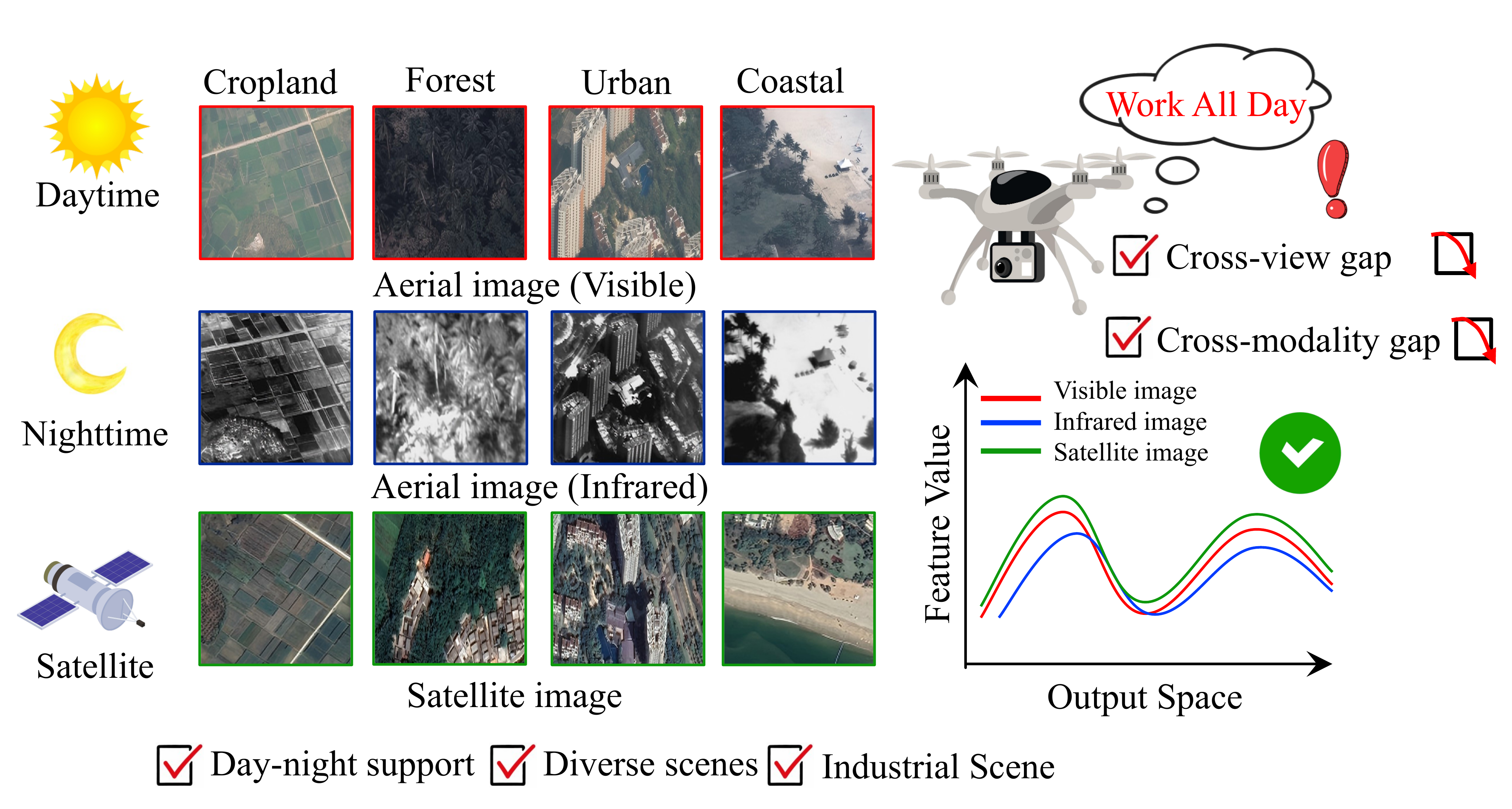}
    \caption{\textbf{Overview of IRVSGeo benchmark.}
    IRVSGeo provides geo-tagged visible aerial, infrared aerial, and satellite reference images across cropland, forest, urban, and coastal scenes. Its aerial images are acquired in real-world conditions, which supports a unified cross-modality evaluation of DVGL methods for day-night tasks. This benchmark facilitates model training and evaluation toward bridging cross-view and cross-modality gaps in such tasks.}
    \label{fig2}
\end{figure}

To address these challenges, IRVSGeo is first introduced, which is a unified benchmark collected in practical environments for DVGL under varying illumination conditions. IRVSGeo contains 26,460 images, which were collected from 8,820 geographic locations using an industrial-grade airborne imaging platform. As shown in Fig.~\ref{fig2}, each location provides one visible aerial image, one infrared aerial image, and one corresponding satellite reference image. The benchmark covers four representative scene categories, which are cropland, forest, urban, and coastal, and enables systematic evaluation under varying illumination conditions via dual sensing modalities at the same geographic locations, with four corresponding subsets divided accordingly. On the basis of this benchmark, the modality-adaptive state-space transport relation network (MASTR-Net) is proposed, which is a multimodal DVGL framework designed to jointly address the appearance discrepancies between visible and infrared modalities and the structural discrepancies between images captured by drone and satellite. Specifically, MASTR-Net employs a modality-adaptive feature enhancement (MAFE) module that uses the global descriptor of each input to guide the modulation of residual updates, thereby adaptively refining spatial feature map responses while preserving the original global descriptor. It further introduces a bidirectional selective state-space relation mixer (BSSRM) to capture long-range structural dependencies and generate relation-aware descriptors for cross-view matching. In addition, a soft optimal transport relation alignment (SOTRA) loss is developed that constructs a balanced soft target from pairwise similarities between the preserved aerial and satellite global descriptors. It aligns the aerial-to-satellite and satellite-to-aerial relation distributions computed from relation-aware descriptors with this target and its transpose, respectively. This loss preserves global matching geometry and provides stable supervision for cross-modality and cross-view relation learning. Extensive experiments demonstrate that MASTR-Net achieves state-of-the-art geo-localization performance on IRVSGeo under varying illumination conditions and remains competitive on the IR-VL328 and CVGL-RGBT infrared benchmarks. The main contributions of this work are summarized as follows.

\begin{itemize}
\item IRVSGeo is established as a unified benchmark for DVGL tasks, which offers geographically aligned visible aerial images, infrared aerial images, and satellite reference images at the same locations. Its aerial images are acquired under real-world conditions, which facilitates model training and systematic evaluation toward bridging cross-view and cross-modality gaps in such tasks. It holds great potential for practical day-night DVGL tasks.

\item The unified framework termed MASTR-Net is proposed to perform DVGL using both visible and infrared observations. It adapts the feature responses to the input modality to alleviate the appearance discrepancies between visible and infrared imagery. The BSSRM module is then employed to capture long-range structural dependencies between images captured by drone and satellite. The SOTRA loss constructs a balanced soft target from global descriptor similarities and uses it to supervise relation learning.

\item Extensive experiments are conducted on IRVSGeo and two public infrared benchmarks, IR-VL328 and CVGL-RGBT. MASTR-Net consistently achieves strong performance on IRVSGeo and competitive generalization on infrared benchmarks.

\end{itemize}

The remainder of this paper is organized as follows. Section \ref{related works} reviews related work. Section \ref{benchmark} introduces the proposed IRVSGeo benchmark and describes its data collection, organization, and evaluation protocol. Section \ref{method} presents MASTR-Net in detail. Section \ref{experiments} reports the experimental results and the corresponding discussions. Finally, the conclusions are outlined in Section \ref{conclusions}.

\section{RELATED WORK}
\label{related works}

\subsection{DVGL Benchmarks}

DVGL has been formulated as a retrieval task in which a query image captured from an aerial viewpoint is matched with a geo-tagged satellite reference image to identify the geographic location of a target region  \cite{zhang2026geo2,11622851}.  Early benchmarks for cross-view geo-localization (CVGL), which were based on ground–satellite pairs, pioneered this research direction and laid the foundation for subsequent research. CVUSA \cite{zhai2017predicting} and CVACT \cite{liu2019lending} established a large-scale evaluation in which ground panoramas were matched with satellite reference images, while VIGOR \cite{zhu2021vigor} relaxed the assumption that a ground query is located at the center of its corresponding satellite reference image. These benchmarks have driven substantial progress in ground-to-satellite geo-localization. However, they have remained limited to ground-level viewpoints and have not captured the altitude and scale variations introduced by aerial flight, which has limited their applicability to aerial scenarios. With the rapid development and widespread deployment of drones, DVGL has emerged as an important research direction distinct from conventional ground-based CVGL \cite{11602083,11655941}. University-1652 \cite{zheng2020university} provided visible aerial, visible ground, and satellite reference images for drone-to-satellite geo-localization. SUES-200 \cite{zhu2023sues} included visible aerial images captured at multiple flight altitudes to evaluate altitude and scale variations. DenseUAV \cite{dai2023vision} introduced dense sampling in real urban environments, while UAV-VisLoc \cite{xu2024uav} increased the diversity of terrain, flight height, and heading. GTA-UAV \cite{ji2025game4loc} and World-UAV \cite{11077664} further considered continuous geographic locations, partial spatial overlap, and one-to-many correspondence. Despite these advances, existing DVGL benchmarks have remained dominated by visible imagery. IR-VL328 \cite{liu2025object} and CVGL-RGBT \cite{zhou2025cdm} extended DVGL to incorporate infrared observations at night or under low-illumination conditions. However, these benchmarks have generally lacked visible aerial images geographically aligned with infrared aerial and satellite reference images at the same locations. This limitation has made it difficult to evaluate DVGL under varying illumination conditions while keeping geographic locations fixed.

\subsection{DVGL Methods on Single-Modality Benchmarks}

Existing DVGL methods have focused mainly on learning discriminative representations in drone and satellite views. Early studies divided feature maps into spatial regions to improve local correspondence. LPN \cite{wang2021each} adopted square-ring partitioning to extract local patterns, while FSRA \cite{dai2021transformer} used a Transformer to identify and align salient regions. MCCG \cite{shen2023mccg} combined a ConvNeXt backbone with multiple classifiers to learn complementary features. More recent methods have increasingly relied on contrastive learning and fine-grained feature alignment. Sample4Geo \cite{deuser2023sample4geo} introduced symmetric contrastive learning with hard-negative sampling, CAMP \cite{wu2024camp} combined position-aware partitioning with contrastive attribute mining, and DAC \cite{xia2024enhancing} improved global and local correspondence through domain alignment and scene consistency. MFRGN \cite{wang2024mfrgn} integrated multi-scale global and local features to improve generalizability across geographic locations. SURFNet \cite{liu2026surfnet} further modeled ground structures through semantic augmentation, positional encoding, and adaptive feature aggregation to reduce the influence of background changes and spatial misalignment. Another line of research has aimed to reduce dependence on paired annotations. CDIKTNet \cite{11622851} learned cross-domain invariant features from limited paired data and transferred such knowledge to unpaired target domains. DMNIL \cite{ChenWithout} adopted self-supervised clustering and memory-based contrastive learning to discover correspondences without predefined image pairs. Iterative rendering \cite{11010141} reconstructed three-dimensional scenes from multiple aerial observations and produced overhead representations to reduce viewpoint differences without paired training. UniABG \cite{chen2026uniabg} combined adversarial view adaptation with graph-based correspondence refinement to improve pseudo-label reliability in unsupervised DVGL.
This reduced reliance on supervised training and improved its applicability when paired data are scarce or unavailable.

Despite progress in representation learning, generalizability across geographic locations, and label-efficient training, most existing methods have primarily been evaluated on visible daytime benchmarks \cite{zheng2020university,zhu2023sues,dai2023vision,ji2025game4loc}. These methods have largely relied on color, texture, and local appearance cues extracted from visible images. These cues become less reliable at night or under low-illumination conditions. Moreover, unsupervised and limited-supervision methods have mainly addressed annotation requirements rather than explicitly tackling the cross-modality gap between visible and infrared observations. Consequently, these methods have primarily supported daytime tasks, while their direct extension to consistent geo-localization across day and night has remained challenging.

\section{IRVSGeo Benchmark}
\label{benchmark}

\begin{table*}[ht]
\centering
\caption{Comparison of DVGL benchmarks in terms of sensing modality and benchmark property.}
\label{tab:dataset_comparison}
\renewcommand{\arraystretch}{1.08}
\setlength{\tabcolsep}{7pt}

\resizebox{\textwidth}{!}{%
\begin{tabular}{lccc|ccccc}
\hline
\multirow{2}{*}{Benchmark}
& \multicolumn{3}{c}{Sensing Modality}
& \multicolumn{5}{c}{Benchmark Property} \\
\cline{2-4}
\cline{5-9}
& visible aerial
& infrared aerial
& Satellite
& Daytime
& Nighttime
& Scene Diversity
& GPS-tag
& Locations \\
\hline

University-1652\cite{zheng2020university}
& \cmark
& \xmark
& \cmark
& \cmark
& \xmark
& \xmark
& \xmark
& 1652 \\

SUES-200\cite{zhu2023sues}
& \cmark
& \xmark
& \cmark
& \cmark
& \xmark
& \cmark
& \xmark
& 200 \\

IR-VL328\cite{liu2025object}
& \xmark
& \cmark
& \cmark
& \xmark
& \cmark
& \xmark
& \xmark
& 328 \\

CVGL-RGBT\cite{zhou2025cdm}
& \xmark
& \cmark
& \cmark
& \xmark
& \cmark
& \cmark
& \xmark
& 748 \\

IRVSGeo (Proposed)
& \cmark
& \cmark
& \cmark
& \cmark
& \cmark
& \cmark
& \cmark
& 8,820 \\

\hline
\end{tabular}%
}
\end{table*}


\subsection{Data Acquisition}

IRVSGeo has been collected using an industrial-grade airborne imaging system over a large region that extends from Wanning to Lingshui in Hainan Province, China. The acquisition area covers tens of kilometers and exhibits substantial variations in land cover, scene structure, and human activity. It includes cropland, rural roads, coastal areas, nearshore waters, forests, residential communities, and urban buildings.

The airborne platform acquires visible and infrared aerial images at geographically referenced locations. Geographic coordinates link each aerial observation with its corresponding satellite reference image. Thus, each geographic location is represented by one visible aerial image, one infrared aerial image, and one satellite reference image. These images are geographically aligned rather than pixel-wise registered because aerial and satellite imagery differ in viewpoint, spatial resolution, and imaging geometry.

\begin{figure}[t]
    \centering
    \includegraphics[width=\columnwidth]{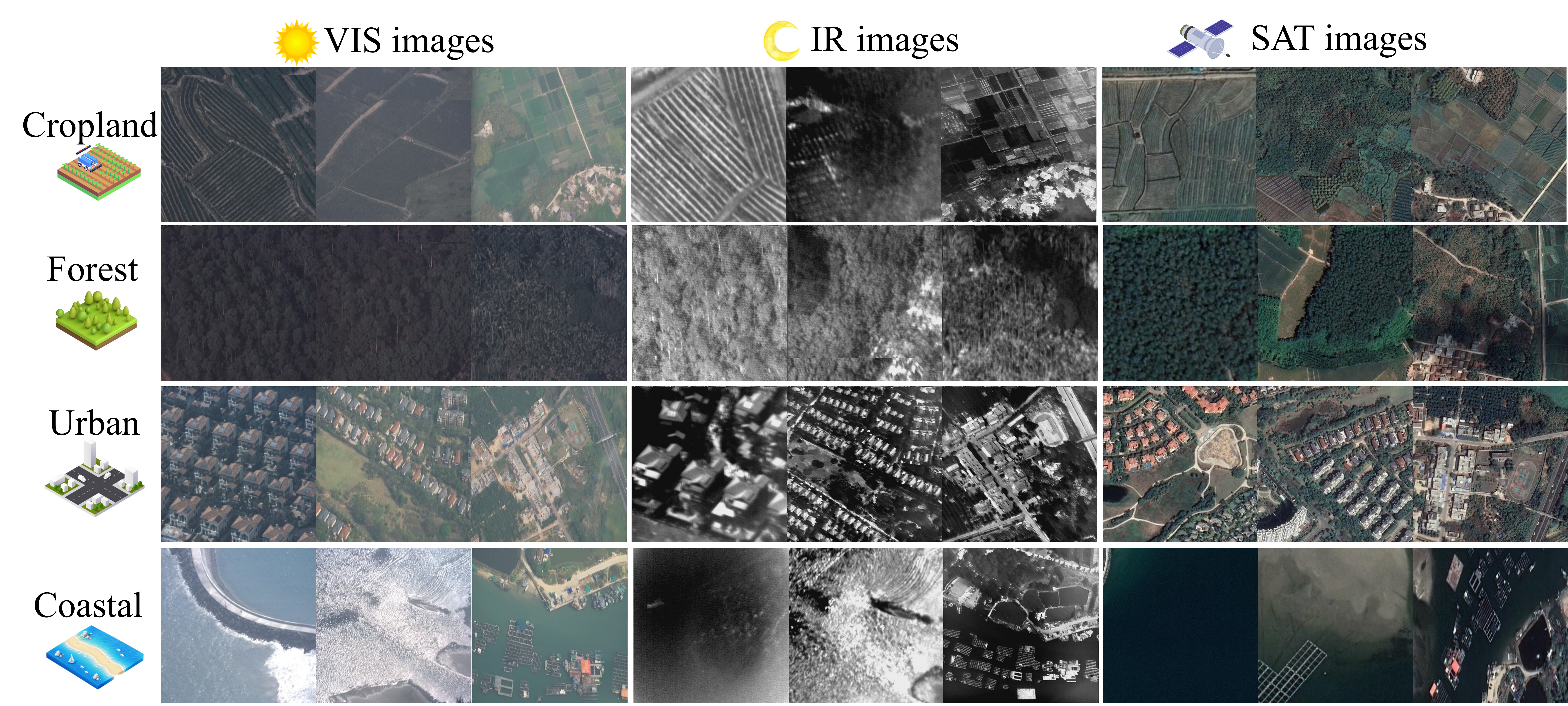}
\caption{\textbf{Illustration of IRVSGeo.}  Each row corresponds to one scene category, while columns display geographically aligned visible aerial, infrared aerial, and satellite reference images. These images share identical location identifiers but exhibit significant differences in viewpoint, spatial resolution, and visual appearance.}
    \label{fig:IRVSGeo_samples}
\end{figure}

Grouped by dominant geographic characteristics, the sampled locations are partitioned into four scene categories: Cropland, Forest, Urban, and Coastal. Representative samples are shown in Fig.~\ref{fig:IRVSGeo_samples}. The four categories exhibit distinctly different visual patterns. Cropland scenes are characterized by repetitive field boundaries and rural roads. Forest scenes are dominated by dense vegetation with weak structural cues. Urban scenes comprise buildings, roads, and residential blocks with complex spatial layouts. Coastal scenes contain irregular shorelines and extensive water regions. 

\begin{table}[t]
\centering
\caption{Data composition and splits of IRVSGeo.}
\label{tab:IRVSGeo_data_composition}
\renewcommand{\arraystretch}{1.08}
\setlength{\tabcolsep}{3pt}

\resizebox{\columnwidth}{!}{%
\begin{tabular}{cccccc}
\hline
\multicolumn{2}{c}{\multirow[c]{2}{*}{Split}}
& \multicolumn{3}{c}{Number of Images}
& \multirow[c]{2}{*}{\makecell{Number of\\Locations}} \\
\cline{3-5}
\multicolumn{2}{c}{}
& visible aerial
& infrared aerial
& Satellite
& \\
\hline

\multicolumn{2}{c}{Training set}
& 7,056
& 7,056
& 7,056
& 7,056 \\

\multirow[c]{2}{*}{Test set}
& $\mathcal{A}\rightarrow\mathcal{S}$
& 882
& 882
& 882
& 882 \\

& $\mathcal{S}\rightarrow\mathcal{A}$
& 882
& 882
& 882
& 882 \\

\hline
\multicolumn{2}{c}{Total}
& 8,820
& 8,820
& 8,820
& 8,820 \\
\hline
\end{tabular}%
}
\end{table}

\subsection{Benchmark Composition and Splits}

IRVSGeo contains 26,460 images from 8,820 geographic locations. Each location provides one visible aerial image, one infrared aerial image, and one satellite reference image, resulting in 8,820 images for each modality.

Table~\ref{tab:dataset_comparison} compares IRVSGeo with representative DVGL benchmarks in terms of sensing modalities and benchmark properties.

As reported in Table~\ref{tab:IRVSGeo_data_composition}, 7,056 locations with 21,168 images are used for training, termed the training set. The remaining 1,764 locations with 5,292 images are reserved for evaluation, termed the test set. The test set is divided into two non-overlapping groups based on the retrieval direction. One group contains 882 locations for geo-localization in which visible or infrared aerial images are used as queries and satellite reference images are used as the gallery, denoted as $\mathcal{A}\rightarrow\mathcal{S}$. The other group contains 882 different locations for the reverse direction, where satellite reference images are used as queries, and visible or infrared aerial images serve as the gallery, denoted as $\mathcal{S}\rightarrow\mathcal{A}$. Within each retrieval direction, query and gallery images are associated by the same geographic location identity. Notably, no location is shared between the two retrieval directions; that is, each geographic location is assigned exclusively to
one retrieval direction.

As shown in Table~\ref{tab:IRVSGeo_scene_statistics}, the training set contains 1,516 Cropland locations, 1,268 Coastal locations, 3,448 Forest locations, and 824 Urban locations. Each retrieval direction in the test set includes 190 Cropland locations, 158 Coastal locations, 431 Forest locations, and 103 Urban locations. The relative proportions of the four scene categories are preserved in all training and test sets. Since each location contains both a visible and an infrared aerial image, the count of aerial images within each scene category is twice the number of satellite reference images.

\begin{table}[!t]
\centering
\caption{IRVSGeo subset statistics. Aerial counts include both visible and infrared aerial images.}
\label{tab:IRVSGeo_scene_statistics}
\renewcommand{\arraystretch}{1.08}
\setlength{\tabcolsep}{3pt}

\resizebox{\columnwidth}{!}{%
\begin{tabular}{ccccccc}
\hline
\multirow[c]{2}{*}{Subset}
& \multicolumn{2}{c}{Training set}
& \multicolumn{2}{c}{Test set $\mathcal{A}\rightarrow\mathcal{S}$}
& \multicolumn{2}{c}{Test set $\mathcal{S}\rightarrow\mathcal{A}$} \\
\cline{2-3}
\cline{4-5}
\cline{6-7}
& Aerial
& Satellite
& Aerial
& Satellite
& Aerial
& Satellite \\
\hline

Cropland
& 3,032
& 1,516
& 380
& 190
& 380
& 190 \\

Coastal
& 2,536
& 1,268
& 316
& 158
& 316
& 158 \\

Forest
& 6,896
& 3,448
& 862
& 431
& 862
& 431 \\

Urban
& 1,648
& 824
& 206
& 103
& 206
& 103 \\

\hline
Total
& 14,112
& 7,056
& 1,764
& 882
& 1,764
& 882 \\
\hline
\end{tabular}%
}
\end{table}

\subsection{Real-World Characteristics}

IRVSGeo retains visible aerial images affected by degradations encountered during real-world acquisition, without discarding low-quality samples through strict manual filtering. Visible aerial images suffer from overexposure, low-illumination, strong shadows, occlusion, motion blur, defocus, and limited spatial resolution. Variations in aircraft attitude also produce irregular oblique viewpoints, which increase the cross-view structural discrepancies between visible aerial and satellite reference images. Infrared observations introduce additional challenges. Compared with visible aerial images, infrared aerial images generally contain weaker textures, less distinct boundaries, and less diversity of appearance. Objects with similar thermal responses become difficult to discriminate. Meanwhile, roads, vegetation, water, and buildings can exhibit substantially different appearances in visible and infrared modalities {\cite{cui2021cross,zhou2025cdm}}. The task therefore involves not only the structural discrepancies between images captured by drone and satellite, but also the cross-modality gaps between visible and infrared observations.

The four scene categories introduce different forms of ambiguity. Repeated field patterns render the Cropland scene difficult to distinguish, while large homogeneous water regions reduce discriminative information in Coastal scenes. Forest scenes contain repetitive vegetation textures and weak structural boundaries. Urban scenes exhibit dense objects, occlusion, and substantial changes in scale and orientation. These characteristics make IRVSGeo suitable for the evaluation of DVGL methods under various real-world conditions.

\subsection{Evaluation Protocol} 

As shown in Table \ref{tab:IRVSGeo_data_composition}, two retrieval directions are given, namely $\mathcal{S}\rightarrow\mathcal{A}$ and $\mathcal{A}\rightarrow\mathcal{S}$. Consequently, the following evaluation protocol defines six fine-grained settings for training and testing. 
\begin{enumerate}
\item  $\mathcal{V}\rightarrow\mathcal{S}$ for geo-localization from visible aerial images to satellite reference images, 
\item $\mathcal{I}\rightarrow\mathcal{S}$ for geo-localization from infrared aerial images to satellite reference images, 
\item $\mathcal{V}\oplus\mathcal{I}\rightarrow\mathcal{S}$ for geo-localization from joint visible–infrared aerial images to satellite reference images, 
\item $\mathcal{S}\rightarrow\mathcal{V}$ for geo-localization from satellite reference images to visible aerial images,
\item $\mathcal{S}\rightarrow\mathcal{I}$ for geo-localization from satellite reference images to infrared aerial images,
\item $\mathcal{S}\rightarrow\mathcal{V}\oplus\mathcal{I}$ for geo-localization from satellite reference images to joint visible–infrared aerial images.
\end{enumerate}
Notably, each geographic location is assigned exclusively to one retrieval direction. That is, the sets of geographic locations adopted for the two directions are disjoint.

Among these six settings, only four unimodal-to-unimodal retrieval tasks are evaluated during testing. The query-side fusion setting (\(\mathcal{V}\oplus\mathcal{I}\rightarrow\mathcal{S}\)) and the gallery-side fusion setting (\(\mathcal{S}\rightarrow\mathcal{V}\oplus\mathcal{I}\) ) are used only for training and are not evaluated during testing. Each location contains one visible aerial image, one infrared aerial image, and one satellite reference image. A retrieved image is considered a correct match if it shares the same geographic location as the query. Each query has one positive gallery image in the settings of $\mathcal{V}\rightarrow\mathcal{S}$, $\mathcal{I}\rightarrow\mathcal{S}$, and $\mathcal{V}\oplus\mathcal{I}\rightarrow\mathcal{S}$. For the other three settings, each satellite query has two positive aerial images within the gallery, i.e., the visible and infrared aerial images from the same geographic location. 

For each retrieval direction, we report overall retrieval performance for unimodal query-gallery matching across all locations assigned to that direction and scene-specific performance for Cropland, Coastal, Forest, and Urban. Recall@$K$ and average precision (AP) are adopted as evaluation metrics. Recall@$K$ indicates whether any correctly matched sample appears within the top-$K$ retrieved candidates, while AP quantifies the overall ranking quality. Recall@1 is used to assess the accuracy of the top-ranked retrieved results.

\begin{figure*}[t]
  \centering
  \includegraphics[width=\textwidth]{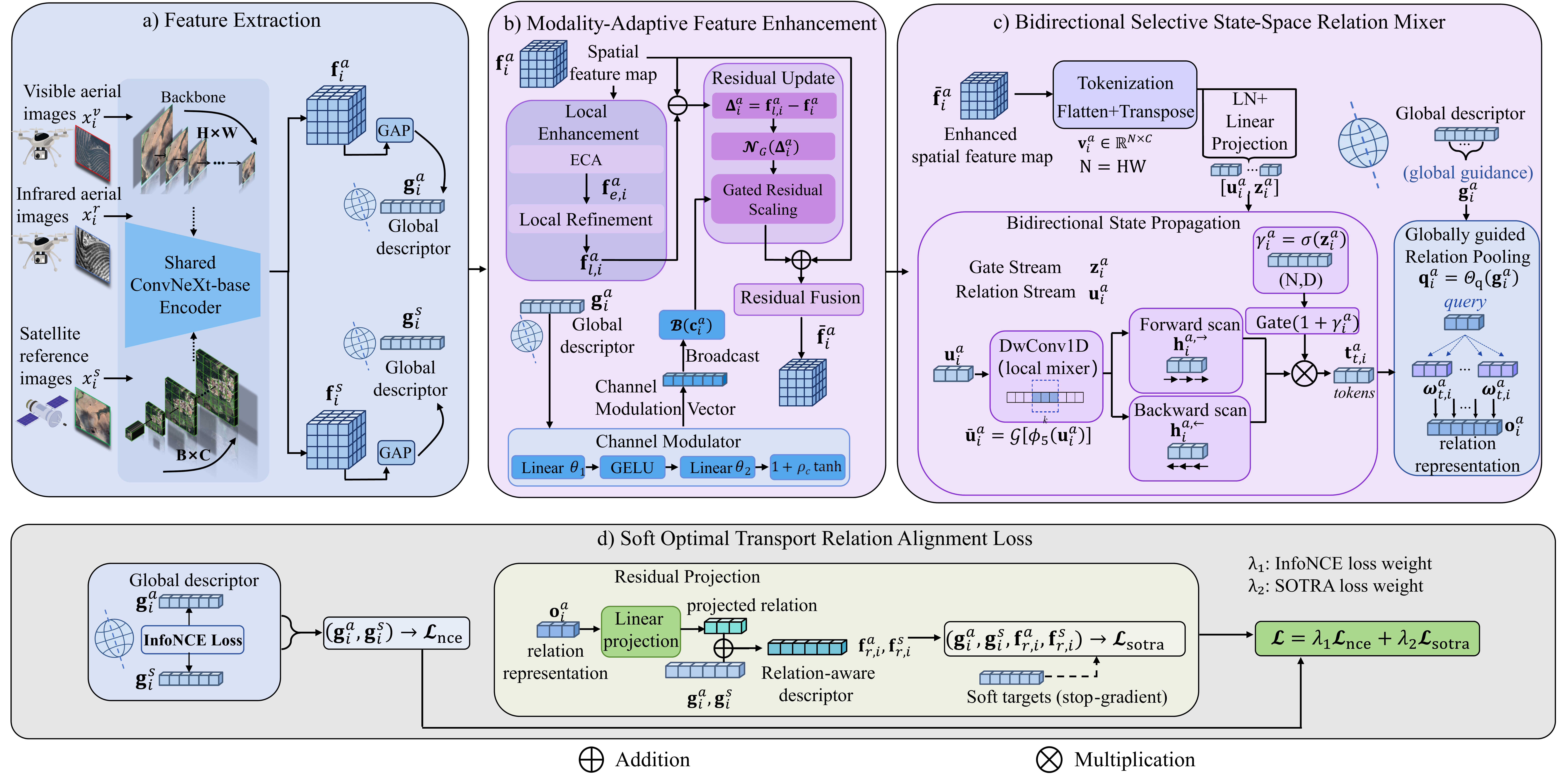}
  \caption{\textbf{Overall pipeline of MASTR-Net.}
(a) The shared encoder extracts spatial feature maps
\(\mathbf f_i^{a},\mathbf f_i^{s}\) and global descriptors
\(\mathbf g_i^{a},\mathbf g_i^{s}\).
(b) MAFE enhances aerial spatial feature maps to obtain
\(\bar{\mathbf f}_i^{a}\).
(c) BSSRM captures bidirectional dependencies and produces
relation representations \(\mathbf o_i^{a}\) under global guidance.
(d) Residual projection generates
\(\mathbf f_{r,i}^{a},\mathbf f_{r,i}^{s}\);
InfoNCE supervises global descriptors, while SOTRA supervises
relation-aware descriptors using soft targets derived from
global descriptors.
}
  \label{fig4}
\end{figure*}

\section{Modality-Adaptive State-Space Transport Relation Network}
\label{method}

\subsection{Problem Formulation} 

Given a visible aerial image set $\mathcal{X}^{v}=\{x_i^{v}\}_{i=1}^{N}$, an infrared aerial image set $\mathcal{X}^{r}=\{x_i^{r}\}_{i=1}^{N}$, and a satellite reference image set $\mathcal{X}^{s}=\{x_i^{s}\}_{i=1}^{N}$, the images $x_i^{v}$, $x_i^{r}$, and $x_i^{s}$ correspond to the same geographic location, where $N$ is the number of geographic locations. During training, $x_i^{v}$, $x_i^{r}$, or their joint representation is selected as the drone-view query and paired with $x_i^{s}$ as its positive sample, whereas images from different geographic locations are treated as negative samples. The objective is to learn a shared embedding space that brings aerial and satellite reference images from the same geographic location close together while separating images from different locations. Unlike visible-only DVGL settings, our protocol incorporates both visible and infrared aerial images. These aerial images serve as queries in the retrieval direction  $\mathcal{A}\rightarrow\mathcal{S}$ and as galleries in the retrieval direction $\mathcal{S}\rightarrow\mathcal{A}$, enabling performance evaluation across both modalities and retrieval directions.

\subsection{Shared ConvNeXt-based Encoder}

For notational simplicity, an aerial image is denoted by $x_i^{a}$, where $x_i^{a}$ can be a visible aerial image $x_i^{v}$ or an infrared aerial image $x_i^{r}$. Visible imagery provides rich appearance cues under favorable illumination, whereas infrared imagery remains informative at night or under low-illumination conditions. The same geo-localization framework can therefore process either modality to support day-night DVGL tasks.

Given an aerial-satellite pair $(x_i^{a}, x_i^{s})$, a shared ConvNeXt-base encoder \cite{liu2022convnet} is used to extract their feature maps,
\begin{equation}
\mathbf{f}_i^{a}
=
\mathcal{F}_{\mathrm{b}}(x_i^{a}),
\qquad
\mathbf{f}_i^{s}
=
\mathcal{F}_{\mathrm{b}}(x_i^{s}),
\label{eq:backbone}
\end{equation}
where $\mathcal{F}_{\mathrm{b}}$ denotes the shared ConvNeXt-base encoder, and $\mathbf{f}_i^{a},\mathbf{f}_i^{s}
\in\mathbb{R}^{C\times H\times W}$ denote the aerial and satellite spatial feature maps, respectively. Global average pooling
$\mathcal{P}_{\mathrm{a}}(\cdot)$ over the spatial dimensions yields the global descriptors
$\mathbf{g}_i^{a}=\mathcal{P}_{\mathrm{a}}(\mathbf{f}_i^{a})
\in\mathbb{R}^{C}$ and
$\mathbf{g}_i^{s}=\mathcal{P}_{\mathrm{a}}(\mathbf{f}_i^{s})
\in\mathbb{R}^{C}$.

\subsection{Modality-Balanced Joint Training}

Although IRVSGeo provides geographically aligned triplets $\{x_i^{v}, x_i^{r}, x_i^{s}\}$, if visible aerial and infrared aerial images from the same location are directly included within a single batch, duplicate geographic samples would appear in this batch, which violates the one-to-one matching assumption underlying the subsequent contrastive and relational alignment objectives. Therefore, a modality-balanced joint training strategy is adopted. For a batch of size $B$, each geographic location appears only once per batch, paired with its satellite reference image and either its visible aerial image or its infrared aerial image. The visible and infrared pairs each account for $B/2$ samples, and their geographic locations do not overlap within the same batch. Within each batch, half of the samples are visible imagery and the other half are infrared imagery, with sample indices randomly permuted, which allows the model to receive balanced supervision from visible imagery and infrared imagery while a square pairwise correspondence matrix is preserved.

This training strategy is referred to as modality-balanced joint training because the shared backbone is jointly optimized using data from the two aerial modalities at the data and training levels, rather than via feature-level concatenation or aggregation. Visible aerial, infrared aerial, and satellite reference images are processed by the same MASTR-Net with shared parameters, and no additional fusion head or separate modality-specific inference network is used.

\subsection{Modality-Adaptive Feature Enhancement}

The shared ConvNeXt-base encoder is used to process visible aerial, infrared aerial, and satellite reference images within a common representation space. However, their spatial feature responses may differ substantially because of variations in sensing modality, viewpoint, and illumination. These differences are particularly evident for infrared imagery, in which color information is unavailable and fine-grained textures are often weakened. Since the global descriptor is directly used for retrieval, its modification may alter the image-level semantics learned by the shared encoder. The MAFE module is therefore introduced to refine spatial feature responses and preserve the global descriptor.

{\bf Local enhancement block}. For clarity, the MAFE module is described with the aerial spatial feature map $\mathbf{f}_i^{a}$ as an example. The same transformation is applied to the satellite spatial feature map, with MAFE module parameters shared across aerial and satellite inputs. As illustrated in Fig.~\ref{fig4}(b), the local enhancement block first employs efficient channel attention (ECA) \cite{wang2020eca} to recalibrate the channel responses of $\mathbf{f}_i^{a}$. Specifically, a one-dimensional convolution $\phi_{\mathrm{eca}}^{1\mathrm{D}}(\cdot)$ models local interactions between neighboring channels of the global descriptor $\mathbf{g}_i^{a}$, and a sigmoid function $\sigma(\cdot)$ generates the corresponding channel weights. These weights are broadcast over the spatial dimensions and applied to the original spatial feature map:
\begin{equation}
\begin{aligned}
\mathbf{f}_{\mathrm{e},{i}}^{a}
&=
\mathbf{f}_i^{a}
\odot
\mathcal{B}\!\left[
\sigma\!\left(
\phi_{\mathrm{eca}}^{1\mathrm{D}}
\left(\mathbf{g}_i^{a}\right)
\right)
\right].
\end{aligned}
\label{eq:mafe_eca}
\end{equation}
Here, $\mathcal{B}(\cdot)$ denotes an operation in which a channel vector is broadcast over the spatial dimensions, and $\odot$ denotes element-wise multiplication. This operation recalibrates the channel responses before the following local refinement.

The recalibrated spatial feature map $\mathbf{f}_{\mathrm{e},i}^{a}$ is subsequently processed by a lightweight local refinement in the local enhancement block. A $3\times3$ depthwise convolution $\phi_{3\times3}^{\mathrm{dw}}(\cdot)$ captures local neighborhood information within each channel. The depthwise convolution output is passed through a batch normalization layer and a GELU activation. Then, a \(1\times1\) convolution is applied for cross-channel interaction, followed by a final batch normalization layer. The resulting locally enhanced spatial feature map is defined as
\begin{equation}
\mathbf{f}_{l,i}^{a}
=
\mathrm{BN}\!\left(
\phi_{1\times1}\!\left(
\mathrm{GELU}\!\left(
\mathrm{BN}\!\left(
\phi_{3\times3}^{\mathrm{dw}}
(\mathbf{f}_{\mathrm{e},i}^{a})
\right)
\right)
\right)
\right).
\label{eq:local_refine}
\end{equation}

{\bf Residual update block}. 
The residual update block first computes $\Delta_i^{a}=\mathbf{f}_{l,i}^{a}-\mathbf{f}_i^{a}$, where $\mathbf{f}_{l,i}^{a}$ is the locally enhanced spatial feature map produced by the local refinement and $\Delta_i^{a}$ denotes the residual update for the aerial spatial feature map. The preserved global descriptor $\mathbf{g}_i^{a}$ guides the gated residual scaling operation through a channel modulation vector $\mathbf{c}_i^{a}\in\mathbb{R}^{C}$, which is generated by the channel modulator block described below. The gated residual update is computed as
\begin{equation}
\Delta_{\mathrm{g},i}^{a}
=
\tanh(\alpha_m)\,
\mathcal{B}(\mathbf{c}_i^{a})
\odot
\mathcal{N}_{G}(\Delta_i^{a}),
\end{equation}
where $\Delta_{\mathrm{g},i}^{a}$ denotes the gated residual update, $\mathcal{N}_{G}$ denotes group normalization, $\mathcal{B}$ broadcasts the channel modulation vector
over the spatial dimensions, and $\odot$ denotes element-wise multiplication. The learnable scalar $\alpha_m$ controls the residual magnitude through its bounded transformation
$\tanh(\alpha_m)$. This scaling is applied after group normalization, allowing it to directly control the magnitude of the normalized residual update.

{\bf Channel modulator block}.
The channel modulator generates $\mathbf{c}_i^{a}$ from the preserved global descriptor $\mathbf{g}_i^{a}$. Specifically, the first projection $\theta_1(\cdot)$ maps $\mathbf{g}_i^{a}$ into a hidden space, followed by the GELU activation. The second projection $\theta_2(\cdot)$ maps the response back to the original channel dimension. The channel modulation vector is computed as
\begin{equation}
\begin{aligned}
\mathbf{c}_i^{a}
&=
\mathbf{1}
+
\rho_c\,
\tanh\!\Bigl(
\theta_2\!\bigl(
\operatorname{GELU}\!\bigl(
\theta_1(\mathbf{g}_i^{a})
\bigr)
\bigr)
\Bigr),
\end{aligned}
\label{eq:mafe_channel}
\end{equation}
where $\mathbf{1}\in\mathbb{R}^{C}$ denotes an all-one vector and $\rho_c$ controls the modulation range. We set $\rho_c=0.10$, so each element of $\mathbf{c}_i^{a}$
lies in $(1-\rho_c,1+\rho_c)=(0.9,1.1)$, providing bounded channel modulation of the residual update. The parameters of $\theta_2(\cdot)$ are initialized to zero,
so $\mathbf{c}_i^{a}$ is initially an all-one vector, yielding identity channel modulation at initialization.

{\bf Residual fusion block}.
The residual fusion block performs residual addition by adding the gated residual update $\Delta_{\mathrm{g},i}^{a}$ to the original spatial feature map $\mathbf{f}_i^{a}$:
\begin{equation}
\bar{\mathbf{f}}_i^{a}
=
\mathbf{f}_i^{a}
+
\Delta_{\mathrm{g},i}^{a},
\end{equation}
where $\bar{\mathbf{f}}_i^{a}$ denotes the enhanced aerial spatial feature map. In this manner, the MAFE module performs bounded, input-dependent modulation of the spatial feature map
while the preserved global descriptor remains unchanged.

Applying the shared MAFE module to the satellite spatial feature map similarly produces $\bar{\mathbf{f}}_i^{s}$. The enhanced spatial feature maps $\bar{\mathbf{f}}_i^{a}$ and $\bar{\mathbf{f}}_i^{s}$ are subsequently passed to the BSSRM module for relation modeling, while the preserved global descriptors $\mathbf{g}_i^{a}$ and $\mathbf{g}_i^{s}$ provide global
semantic guidance for globally guided relation pooling, retrieval, and the construction of the balanced soft target.

\subsection{Bidirectional Selective State-Space Relation Mixer}

Although the MAFE module adapts local spatial responses to different input modalities, it does not explicitly capture dependencies between distant regions. Under large viewpoint and scale variations, corresponding structures may appear at different spatial positions in aerial and satellite reference images. Meanwhile, different locations may contain similar local patterns. The MAFE module is therefore insufficient to distinguish challenging cross-view candidates. Therefore, the BSSRM module is introduced, which combines local context, bidirectional state propagation, and global semantic guidance to construct relation-aware descriptors.

For clarity, we describe the BSSRM module using aerial features as an example. Satellite features are processed in the same way with shared BSSRM parameters.

{\bf Bidirectional state propagation block}.
As illustrated in Fig.~\ref{fig4}(c), BSSRM models relations among spatial locations using the enhanced spatial feature map $\bar{\mathbf{f}}_i^{a}\in\mathbb{R}^{C\times H\times W}$.
The preserved global descriptor $\mathbf{g}_i^{a}\in\mathbb{R}^{C}$ is retained to provide global semantic guidance for subsequent relation pooling. To prepare the input sequence for bidirectional state propagation, we flatten the spatial dimensions of $\bar{\mathbf{f}}_i^{a}$ and transpose the resulting tensor to obtain $\mathbf{v}_i^{a}=\mathcal{R}(\bar{\mathbf{f}}_i^{a})\in\mathbb{R}^{N\times C}$, where $N=HW$ and $\mathcal{R}$ denotes this rearrangement. Each token corresponds to one spatial location and retains its $C$-dimensional feature vector. We then apply layer normalization to each token along the channel dimension, followed by a linear projection from $C$ to $2D$ dimensions. The projected features are split into two $D$-dimensional streams: a relation stream that supplies the content for bidirectional state propagation and a gate stream that adaptively modulates the resulting relation responses:
\begin{equation}
[\mathbf{u}_i^{a},\mathbf{z}_i^{a}]
=
\mathcal{N}_{L}(\mathbf{v}_i^{a})\Theta_{n},
\label{eq:bssrm_projection}
\end{equation}
where $\mathbf{u}_i^{a},\mathbf{z}_i^{a}\in\mathbb{R}^{N\times D}$ denote the relation and gate streams, respectively.

The resulting relation and gate streams are further processed separately. The relation stream is processed by a depthwise convolution and GELU activation, which incorporate local neighborhood information and produce the locally mixed relation stream $\bar{\mathbf{u}}_i^{a}$.
In parallel, the gate stream is passed through a sigmoid function to generate the modulation tensor $\boldsymbol{\gamma}_i^{a}$.
The two streams are formulated as
\begin{equation}
\left(
\bar{\mathbf{u}}_i^{a},
\boldsymbol{\gamma}_i^{a}
\right)
=
\left(
\mathcal{G}\!\left[
\phi_{5}\!\left(\mathbf{u}_i^{a}\right)
\right],
\,
\sigma\!\left[\mathbf{z}_i^{a}\right]
\right),
\label{eq:bssrm_input}
\end{equation}
where $\phi_5(\cdot)$ denotes a depthwise convolution with a kernel size of 5 along the token dimension. Both $\bar{\mathbf{u}}_i^{a}$ and $\boldsymbol{\gamma}_i^{a}$ belong to $\mathbb{R}^{N\times D}$. The former carries locally mixed structural information for subsequent bidirectional state propagation, whereas the latter controls the contribution of individual token-channel responses during bidirectional fusion. The local mixing operation captures only short-range context. The BSSRM module therefore propagates $\bar{\mathbf{u}}_i^{a}$ in both forward and backward directions to model long-range structural dependencies \cite{zhu2024visionmamba}. Let $o\in\{f,b\}$ denote the scanning direction, where $f$ and $b$ correspond to the forward and backward hidden-state sequences $\mathbf{h}_i^{a,\rightarrow}$ and $\mathbf{h}_i^{a,\leftarrow}$ shown in Fig.~\ref{fig4}(c), respectively. The directional input sequences are defined as $\bar{\mathbf{u}}_{t,i}^{a,f}=\bar{\mathbf{u}}_{t,i}^{a}$ and $\bar{\mathbf{u}}_{t,i}^{a,b}=\bar{\mathbf{u}}_{N+1-t,i}^{a}$. Here, $t=1,\ldots,N$ indexes the scanning steps within each directional sequence.

For each scanning direction, an input-dependent update gate is computed as
$\boldsymbol{\beta}_{t,i}^{o}
=\sigma(\bar{\mathbf{u}}_{t,i}^{a,o}\Theta_{\beta})$,
where $\Theta_{\beta}$ denotes a learnable projection. A learnable base-retention vector $\boldsymbol{\mu}^{o}
=\sigma(\boldsymbol{\alpha}^{o})$
controls the persistent memory of each hidden channel.
The resulting retention factor is
$\boldsymbol{\eta}_{t,i}^{o}
=\mathbf{1}
-\boldsymbol{\beta}_{t,i}^{o}
\odot(\mathbf{1}-\boldsymbol{\mu}^{o})$.
The hidden state is updated by
\begin{equation}
\mathbf{h}_{t,i}^{a,o}
=
\boldsymbol{\eta}_{t,i}^{o}
\odot
\mathbf{h}_{t-1,i}^{a,o}
+
\boldsymbol{\beta}_{t,i}^{o}
\odot
\bar{\mathbf{u}}_{t,i}^{a,o},
\label{eq:bssrm_scan}
\end{equation}
where $\mathbf{h}_{0,i}^{a,o}=\mathbf{0}$.
The retention factor determines how much historical structural
information is preserved, while the update gate controls
the contribution of the current token. After the two scans, the backward hidden states are restored to the
original token order and averaged with the forward hidden states. The
modulation tensor $\boldsymbol{\gamma}_{t,i}^{a}$ then modulates the fused response
to produce the relation tokens:
\begin{equation}
\mathbf{t}_{t,i}^{a}
=
\frac{1}{2}\left(
\mathbf{h}_{t,i}^{a,\rightarrow}
+
\mathbf{h}_{N+1-t,i}^{a,\leftarrow}
\right)
\odot
\left(
\mathbf{1}
+
\boldsymbol{\gamma}_{t,i}^{a}
\right),
\label{eq:bssrm_fusion}
\end{equation}
where $\mathbf{t}_{t,i}^{a}\in\mathbb{R}^{D}$.
The bidirectional fusion incorporates structural context from both
scanning directions, while $\boldsymbol{\gamma}_{t,i}^{a}$ adaptively
modulates the response of each token and hidden channel. The relation tokens encode long-range structural dependencies, but
their contributions to location discrimination are not equally
important.

{\bf Globally guided relation pooling}.
The BSSRM module therefore uses the preserved global descriptor
$\mathbf{g}_i^{a}$ to perform globally guided relation pooling.
It is projected into the relation space to form the global query
$\mathbf{q}_i^{a}=\theta_q(\mathbf{g}_i^{a})$,
where $\theta_q(\cdot)$ denotes a learnable linear projection
from $\mathbb{R}^{C}$ to $\mathbb{R}^{D}$.
The weights for globally guided relation pooling are computed as
\begin{equation}
\omega_{t,i}^{a}
=
\operatorname{softmax}_{t}
\left(
\frac{
\left\langle
\mathcal{N}_{\mathrm{D}}\left(\mathbf{t}_{t,i}^{a}\right),
\mathbf{q}_i^{a}
\right\rangle
}{
\sqrt{D}
}
\right),
\label{eq:bssrm_weight}
\end{equation}
where $\omega_{t,i}^{a}$ denotes the normalized scalar weight
assigned to the $t$-th relation token of the $i$-th image
during globally guided relation pooling.
$\mathcal{N}_{\mathrm{D}}(\cdot)$ denotes normalization along
the hidden dimension, and $\operatorname{softmax}_{t}(\cdot)$
normalizes the relevance scores over all relation tokens
within the $i$-th image.

These weights adaptively measure the semantic relevance between
each relation token and the global descriptor.
The globally guided relation representation is generated
as a weighted sum of all relation tokens according to these weights:
\begin{equation}
\mathbf{o}_i^{a}
=
\sum_{t=1}^{N}
\omega_{t,i}^{a}\mathbf{t}_{t,i}^{a},
\label{eq:bssrm_pooling}
\end{equation}
where $\mathbf{o}_i^{a}\in\mathbb{R}^{D}$ denotes the globally
guided relation representation.
This operation emphasizes relation tokens according to
the global semantic guidance provided by the preserved
global descriptor.

\begin{figure}[!t]
    \centering
    \includegraphics[width=\columnwidth]{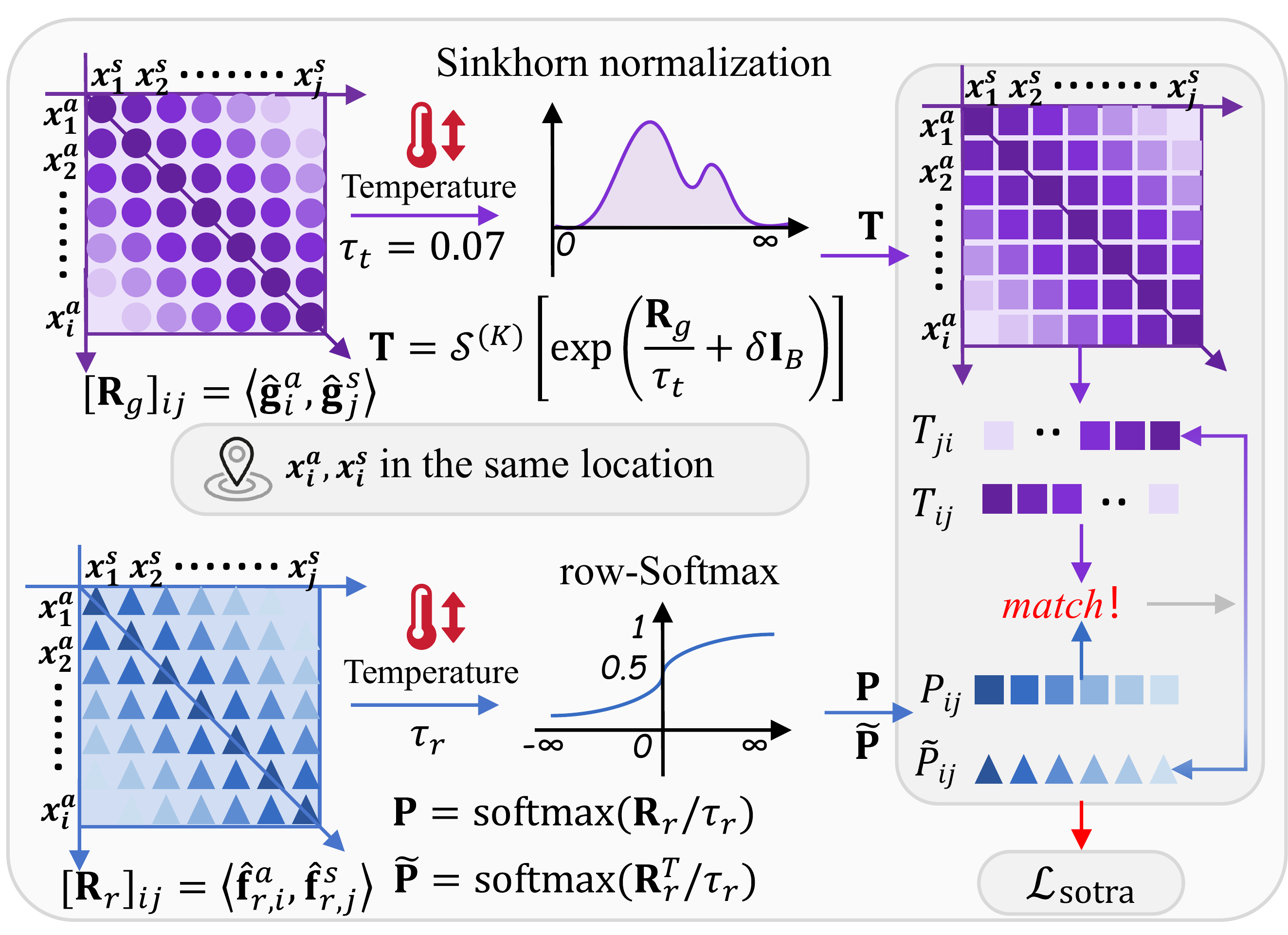}
    \caption{\textbf{Matching and supervision mechanism of SOTRA.}
    The normalized global aerial and satellite descriptors construct
    the global similarity matrix \(\mathbf R_g\).
    After temperature scaling with \(\tau_t\) and the addition of
    the diagonal bias \(\delta\mathbf I_B\), Sinkhorn normalization
    produces the balanced soft target \(\mathbf T\).
    The normalized relation-aware descriptors construct the relation
    similarity matrix \(\mathbf R_r\).
    Row-wise softmax with temperature \(\tau_r\) is applied to
    \(\mathbf R_r\) and its transpose to obtain the aerial-to-satellite
    and satellite-to-aerial distributions \(\mathbf P\) and
    \(\widetilde{\mathbf P}\), respectively.
    For an aerial query \(x_i^a\), \(T_{ij}\) and \(T_{ji}\) provide
    the corresponding forward and reverse matching targets.
    The SOTRA loss \(\mathcal L_{\mathrm{sotra}}\) aligns
    \(\mathbf P\) with \(\mathbf T\) and
    \(\widetilde{\mathbf P}\) with \(\mathbf T^\top\).}
    \label{fig:sotra}
\end{figure}

\subsection{Soft Optimal Transport Relation Alignment Loss}

The residual projection block normalizes $\mathbf{o}_i^{a}$,
projects it back to the original channel dimension, and applies
GELU. The resulting structural relation response is scaled by
a learnable coefficient and added to the preserved global descriptor:
\begin{equation}
\mathbf{f}_{r,i}^{a}
=
\mathcal{N}_{\mathrm{C}}\!\left[
\mathbf{g}_i^{a}
+
\tanh\!\left(\alpha_r\right)
\mathcal{G}\!\left(
\mathcal{N}_{\mathrm{D}}\!\left(\mathbf{o}_i^{a}\right)
\Theta_o
\right)
\right],
\label{eq:bssrm_output}
\end{equation}
where $\Theta_o\in\mathbb{R}^{D\times C}$ denotes the learnable
output projection matrix,
$\mathcal{N}_{\mathrm{C}}(\cdot)$ denotes $\ell_2$ normalization
along the channel dimension, and $\alpha_r$ controls
the contribution of the structural relation response.
The resulting $\mathbf{f}_{r,i}^{a}\in\mathbb{R}^{C}$
serves as the relation-aware aerial descriptor.

Similarly, the relation-aware satellite descriptor
$\mathbf{f}_{r,i}^{s}$ is obtained by applying the same
residual projection to the satellite relation representation
$\mathbf{o}_i^{s}$ and preserved global descriptor
$\mathbf{g}_i^{s}$.

As shown in Fig.~\ref{fig4}(d), residual projection is performed under the associated loss supervision. InfoNCE supervises the global descriptors,
whereas SOTRA supervises bidirectional cross-view matching
distributions across a mini-batch.

The MAFE module retains the global descriptors
$\mathbf{g}_i^{a}$ and $\mathbf{g}_i^{s}$,
while the BSSRM module produces the relation-aware descriptors
$\mathbf{f}_{r,i}^{a}$ and $\mathbf{f}_{r,i}^{s}$. In this formulation, the BSSRM module and SOTRA loss are jointly designed as a coupled relation modeling-and-supervision mechanism: the BSSRM module constructs relation-aware descriptors, whereas SOTRA provides their dedicated training supervision. Specifically, since hard identity labels cannot capture the relative similarity among different locations, SOTRA uses the balanced soft target constructed from the global similarity matrix to supervise relation modeling.

Given a batch of $B$ aerial-satellite pairs, let
$\mathbf{g}_i^{a},\mathbf{g}_i^{s}\in\mathbb{R}^{C}$
denote the global descriptors of the $i$-th pair, and let
$\mathbf{f}_{r,i}^{a},\mathbf{f}_{r,i}^{s}\in\mathbb{R}^{C}$
denote the corresponding relation-aware descriptors,
as illustrated in Fig.~\ref{fig:sotra}.
Each descriptor $\mathbf{x}\in\mathbb{R}^{C}$ is normalized as
$\widehat{\mathbf{x}}=\mathbf{x}/\|\mathbf{x}\|_2$.
The global and relation similarity matrices
$\mathbf{R}_g,\mathbf{R}_r\in\mathbb{R}^{B\times B}$
are defined elementwise as
\[
\begin{aligned}
[\mathbf{R}_g]_{ij}
&=
\left\langle
\widehat{\mathbf{g}}_i^{a},
\widehat{\mathbf{g}}_j^{s}
\right\rangle,\\
[\mathbf{R}_r]_{ij}
&=
\left\langle
\widehat{\mathbf{f}}_{r,i}^{a},
\widehat{\mathbf{f}}_{r,j}^{s}
\right\rangle,
\end{aligned}
\]
where $i,j\in\{1,\ldots,B\}$ index the aerial and satellite
samples, respectively.
The diagonal entries correspond to paired aerial and satellite
images from the same geographic location.

The global similarity matrix is converted into a balanced soft target
through Sinkhorn normalization \cite{cuturi2013sinkhorn}:
\begin{equation}
\mathbf{T}
=
\mathcal{S}^{(K)}
\left[
\exp
\left(
\frac{\mathbf{R}_{g}}{\tau_t}
+
\delta \mathbf{I}_{B}
\right)
\right],
\label{eq:sotra_target}
\end{equation}
where $\mathcal{S}^{(K)}(\cdot)$ denotes $K$ iterations of
alternating row and column normalization, and $\exp(\cdot)$
is applied elementwise.
Here, $\mathbf{I}_B$ is the $B\times B$ identity matrix,
$\tau_t$ controls the sharpness of the balanced soft target,
and $\delta$ controls the diagonal bias toward paired
aerial-satellite samples.
The target $\mathbf{T}$ is treated as a constant when computing
$\mathcal{L}_{\mathrm{sotra}}$, with gradients stopped through
the target-construction branch.

The relation similarity matrix is optimized to match the balanced soft
target. Let
$\mathbf{P}
=
\operatorname{softmax}(\mathbf{R}_{r}/\tau_{r})$
and
$\widetilde{\mathbf{P}}
=
\operatorname{softmax}(\mathbf{R}_{r}^{\top}/\tau_{r})$,
where softmax is applied row-wise and $\tau_{r}$ controls the sharpness
of the bidirectional relation distributions. Accordingly, $\mathbf{P}$
and $\widetilde{\mathbf{P}}$ represent the aerial-to-satellite and
satellite-to-aerial relation distributions, respectively. The SOTRA loss is defined as
\begin{equation}
\mathcal{L}_{\mathrm{sotra}}
=
-\frac{1}{2B}
\sum_{i=1}^{B}
\sum_{j=1}^{B}
\left(
T_{ij}\log P_{ij}
+
T_{ji}\log \widetilde{P}_{ij}
\right).
\label{eq:sotra_loss}
\end{equation}
This symmetric loss aligns the relation-aware descriptors with the
balanced soft target constructed from the global descriptors.

The global descriptors are additionally supervised by the 
InfoNCE loss $\mathcal{L}_{\mathrm{nce}}$ \cite{oord2018representation,deuser2023sample4geo}. The overall objective is
\begin{equation}
\mathcal{L}
=
\lambda_{\mathrm{1}}\mathcal{L}_{\mathrm{nce}}
+
\lambda_{\mathrm{2}}
\mathcal{L}_{\mathrm{sotra}},
\label{eq:overall_loss}
\end{equation}
where \(\lambda_1\) and \(\lambda_2\) balance the contributions of the InfoNCE loss and the SOTRA loss, respectively.

\begin{table*}[t]
\centering
\caption{Comparison with state-of-the-art methods on IRVSGeo under different training settings. Avg. denotes the average over the two retrieval directions.}
\label{tab:IRVSGeo_comparison}
\setlength{\tabcolsep}{4pt}

\resizebox{\textwidth}{!}{%
\begin{tabular}{lccccccccccccc}
\hline
\multirow{2}{*}{Model}
& \multirow{2}{*}{Venue}
& \multicolumn{2}{c}{ $\mathcal{V}\rightarrow\mathcal{S}$}
& \multicolumn{2}{c}{ $\mathcal{S}\rightarrow\mathcal{V}$}
& \multicolumn{2}{c}{Avg.}
& \multicolumn{2}{c}{ $\mathcal{I}\rightarrow\mathcal{S}$}
& \multicolumn{2}{c}{ $\mathcal{S}\rightarrow\mathcal{I}$}
& \multicolumn{2}{c}{Avg.} \\
\cline{3-14}
& & R@1 & AP & R@1 & AP & R@1 & AP
  & R@1 & AP & R@1 & AP & R@1 & AP \\
\hline

Sample4Geo[8]
& ICCV'2023
& 50.68 & 58.68 & 49.77 & 58.39 & 50.23 & 58.54
& 45.46 & 54.00 & 48.30 & 56.17 & 46.88 & 55.09 \\

MEAN[1]
& TGRS'2025
& 52.76 & 60.50 & 52.94 & 60.75 & 52.85 & 60.63
& 48.75 & 56.83 & 50.34 & 58.01 & 49.55 & 57.42 \\

MFRGN[31]
& ACM MM'2024
& 54.08 & 61.78 & 55.44 & 63.27 & 54.76 & 62.53
& 49.66 & 57.21 & 49.54 & 57.24 & 49.60 & 57.23 \\

CAMP[30]
& TGRS'2024
& 53.17 & 62.92 & 52.26 & 60.29 & 52.72 & 61.61
& 48.64 & 56.69 & 49.09 & 56.81 & 48.87 & 56.75 \\

DAC[2]
& TCSVT'2024
& 55.10 & 62.45 & 55.10 & 62.68 & 55.10 & 62.57
& 52.38 & 59.76 & 51.02 & 58.51 & 51.70 & 59.14 \\

SURFNet[32]
& TGRS'2026
& 53.29 & 61.40 & 54.87 & 62.91 & 54.08 & 62.16
& 51.13 & 58.74 & 51.58 & 58.93 & 51.36 & 58.84 \\
\hline

MASTR-Net$^{\dagger}$
& -
& 56.12 & 63.61 & 57.34 & 64.99 & 56.73 & 64.30
& \textbf{53.97} & 61.12 & 54.65 & 61.68 & 54.31 & 61.40 \\

MASTR-Net$^{\ast}$
& -
& \textbf{57.48} & \textbf{64.80}
& \textbf{58.50} & \textbf{65.92}
& \textbf{57.99} & \textbf{65.36}
& 53.96 & \textbf{61.32}
& \textbf{55.21} & \textbf{62.41}
& \textbf{54.59} & \textbf{61.87} \\
\hline
\end{tabular}%
}
\end{table*}

\begin{table}[t]
\centering
\caption{Performance comparison on the Cropland subset of IRVSGeo.}
\label{tab:scene_cropland}
\footnotesize
\setlength{\tabcolsep}{4.0pt}
\renewcommand{\arraystretch}{1.05}
\resizebox{\columnwidth}{!}{%
\begin{tabular}{llcccccccc}
\hline
\multirow{2}{*}{Scene}
& \multirow{2}{*}{Method}
& \multicolumn{2}{c}{ $\mathcal{V}\rightarrow\mathcal{S}$}
& \multicolumn{2}{c}{ $\mathcal{S}\rightarrow\mathcal{V}$}
& \multicolumn{2}{c}{ $\mathcal{I}\rightarrow\mathcal{S}$}
& \multicolumn{2}{c}{ $\mathcal{S}\rightarrow\mathcal{I}$} \\
\cline{3-10}
& & R@1 & AP & R@1 & AP & R@1 & AP & R@1 & AP \\
\hline

\multirow{8}{*}{Cropland}
& Sample4Geo\cite{deuser2023sample4geo}
& 63.68 & 70.08 & 64.74 & 71.05
& 54.21 & 60.53 & 53.68 & 61.28 \\

& MEAN\cite{chen2024multi}
& 63.68 & 69.69 & 62.11 & 69.28
& 55.79 & 61.55 & 58.42 & 65.20 \\

& MFRGN\cite{wang2024mfrgn}
& 67.37 & 73.00 & 62.63 & 69.17
& 55.79 & 61.51 & 57.37 & 64.15 \\

& CAMP\cite{wu2024camp}
& 61.58 & 67.61 & 59.47 & 66.66
& 53.68 & 59.56 & 47.89 & 55.55 \\

& DAC\cite{xia2024enhancing}
& 67.89 & 72.90 & 64.21 & 70.94
& 57.89 & 64.04 & 60.00 & 66.63 \\

& SURFNet\cite{liu2026surfnet}
& 62.11 & 68.68 & 62.11 & 69.02
& 55.26 & 61.43 & 54.21 & 61.43 \\
\hline

& MASTR-Net$^{\dagger}$
& 67.37 & 72.53 & 65.26 & 71.22
& 57.39 & 63.61 & 57.89 & 64.83 \\

& MASTR-Net$^{\ast}$
& 68.42 & 73.27 & 63.68 & 70.49
& 60.53 & 65.70 & 58.95 & 65.45 \\
\hline
\end{tabular}%
}
\end{table}

\begin{table}[t]
\centering
\caption{Performance comparison on the Coastal subset of IRVSGeo.}
\label{tab:scene_coastal}
\footnotesize
\setlength{\tabcolsep}{4.0pt}
\renewcommand{\arraystretch}{1.05}
\resizebox{\columnwidth}{!}{%
\begin{tabular}{llcccccccc}
\hline
\multirow{2}{*}{Scene}
& \multirow{2}{*}{Method}
& \multicolumn{2}{c}{ $\mathcal{V}\rightarrow\mathcal{S}$}
& \multicolumn{2}{c}{ $\mathcal{S}\rightarrow\mathcal{V}$}
& \multicolumn{2}{c}{ $\mathcal{I}\rightarrow\mathcal{S}$}
& \multicolumn{2}{c}{ $\mathcal{S}\rightarrow\mathcal{I}$} \\
\cline{3-10}
& & R@1 & AP & R@1 & AP & R@1 & AP & R@1 & AP \\
\hline

\multirow{8}{*}{Coastal}
& Sample4Geo\cite{deuser2023sample4geo}
& 43.67 & 52.16 & 44.30 & 53.66
& 43.67 & 51.66 & 37.98 & 46.15 \\

& MEAN\cite{chen2024multi}
& 43.67 & 51.84 & 44.30 & 53.46
& 43.67 & 50.34 & 39.87 & 48.36 \\

& MFRGN\cite{wang2024mfrgn}
& 42.41 & 50.86 & 43.67 & 52.88
& 48.10 & 54.56 & 42.41 & 50.12 \\

& CAMP\cite{wu2024camp}
& 41.77 & 50.14 & 44.30 & 52.67
& 46.20 & 53.32 & 38.61 & 47.22 \\

& DAC\cite{xia2024enhancing}
& 43.67 & 52.35 & 41.14 & 51.14
& 46.84 & 54.16 & 41.77 & 50.17 \\

& SURFNet\cite{liu2026surfnet}
& 47.47 & 54.53 & 43.04 & 52.71
& 44.94 & 52.20 & 37.34 & 46.60 \\
\hline

& MASTR-Net$^{\dagger}$
& 48.10 & 55.12 & 46.84 & 55.51
& 47.47 & 54.80 & 40.51 & 49.02 \\

& MASTR-Net$^{\ast}$
& 45.57 & 53.30 & 45.57 & 54.44
& 51.26 & 57.57 & 42.41 & 50.78 \\
\hline
\end{tabular}%
}
\end{table}

\begin{table}[t]
\centering
\caption{Performance comparison on the Forest subset of IRVSGeo.}
\label{tab:scene_forest}
\footnotesize
\setlength{\tabcolsep}{4.0pt}
\renewcommand{\arraystretch}{1.05}
\resizebox{\columnwidth}{!}{%
\begin{tabular}{llcccccccc}
\hline
\multirow{2}{*}{Scene}
& \multirow{2}{*}{Method}
& \multicolumn{2}{c}{ $\mathcal{V}\rightarrow\mathcal{S}$}
& \multicolumn{2}{c}{ $\mathcal{S}\rightarrow\mathcal{V}$}
& \multicolumn{2}{c}{ $\mathcal{I}\rightarrow\mathcal{S}$}
& \multicolumn{2}{c}{ $\mathcal{S}\rightarrow\mathcal{I}$} \\
\cline{3-10}
& & R@1 & AP & R@1 & AP & R@1 & AP & R@1 & AP \\
\hline

\multirow{8}{*}{Forest}
& Sample4Geo\cite{deuser2023sample4geo}
& 57.08 & 64.45 & 55.45 & 62.89
& 51.28 & 58.58 & 58.47 & 64.94 \\

& MEAN\cite{chen2024multi}
& 62.18 & 68.77 & 59.16 & 66.42
& 56.38 & 62.84 & 56.61 & 63.33 \\

& MFRGN\cite{wang2024mfrgn}
& 56.38 & 63.52 & 56.15 & 63.53
& 53.83 & 60.38 & 53.13 & 60.52 \\

& CAMP\cite{wu2024camp}
& 55.92 & 63.34 & 55.68 & 62.74
& 51.28 & 58.26 & 55.45 & 62.88 \\

& DAC\cite{xia2024enhancing}
& 61.72 & 68.56 & 62.88 & 69.42
& 58.24 & 64.51 & 62.41 & 68.57 \\

& SURFNet\cite{liu2026surfnet}
& 59.86 & 66.72 & 58.93 & 66.11
& 53.60 & 60.48 & 56.15 & 63.22 \\
\hline

& MASTR-Net$^{\dagger}$
& 64.97 & 71.14 & 64.50 & 70.88
& 61.02 & 66.84 & 60.32 & 66.90 \\

& MASTR-Net$^{\ast}$
& 64.97 & 71.06 & 65.20 & 71.34
& 62.88 & 68.38 & 62.18 & 68.65 \\
\hline
\end{tabular}%
}
\end{table}

\begin{table}[t]
\centering
\caption{Performance comparison on the Urban subset of IRVSGeo.}
\label{tab:scene_urban}
\footnotesize
\setlength{\tabcolsep}{4.0pt}
\renewcommand{\arraystretch}{1.05}
\resizebox{\columnwidth}{!}{%
\begin{tabular}{llcccccccc}
\hline
\multirow{2}{*}{Scene}
& \multirow{2}{*}{Method}
& \multicolumn{2}{c}{ $\mathcal{V}\rightarrow\mathcal{S}$}
& \multicolumn{2}{c}{ $\mathcal{S}\rightarrow\mathcal{V}$}
& \multicolumn{2}{c}{ $\mathcal{I}\rightarrow\mathcal{S}$}
& \multicolumn{2}{c}{ $\mathcal{S}\rightarrow\mathcal{I}$} \\
\cline{3-10}
& & R@1 & AP & R@1 & AP & R@1 & AP & R@1 & AP \\
\hline

\multirow{8}{*}{Urban}
& Sample4Geo\cite{deuser2023sample4geo}
& 66.02 & 71.76 & 60.19 & 66.58
& 55.34 & 60.44 & 54.37 & 60.89 \\

& MEAN\cite{chen2024multi}
& 66.99 & 71.74 & 56.31 & 63.72
& 53.40 & 58.45 & 55.38 & 61.78 \\

& MFRGN\cite{wang2024mfrgn}
& 66.02 & 70.74 & 59.22 & 65.91
& 57.28 & 61.51 & 54.37 & 60.36 \\

& CAMP\cite{wu2024camp}
& 61.16 & 67.65 & 56.31 & 63.70
& 56.31 & 60.38 & 51.46 & 58.83 \\

& DAC\cite{xia2024enhancing}
& 63.11 & 68.97 & 57.28 & 64.39
& 61.17 & 65.60 & 56.31 & 63.15 \\

& SURFNet\cite{liu2026surfnet}
& 65.05 & 70.83 & 55.34 & 63.69
& 61.16 & 65.38 & 56.31 & 63.08 \\
\hline

& MASTR-Net$^{\dagger}$
& 69.90 & 74.27 & 57.28 & 64.21
& 59.22 & 63.12 & 55.34 & 61.96 \\

& MASTR-Net$^{\ast}$
& 66.99 & 71.76 & 56.31 & 63.62
& 63.11 & 67.14 & 60.19 & 66.41 \\
\hline
\end{tabular}%
}
\end{table}

\begin{table*}[t]
\centering
\footnotesize
\caption{Ablation study of the proposed components on IRVSGeo.}
\label{tab:component_ablation}
\setlength{\tabcolsep}{3.2pt}
\renewcommand{\arraystretch}{1.10}

\resizebox{\textwidth}{!}{%
\begin{tabular}{@{}ccc cccccccc cccccccc@{}}
\hline

\multicolumn{3}{c}{\multirow{2}{*}{Components}}
& \multicolumn{8}{c}{MASTR-Net$^{\dagger}$}
& \multicolumn{8}{c}{MASTR-Net$^{\ast}$} \\
\cline{4-19}

& &
& \multicolumn{2}{c}{ $\mathcal{V}\rightarrow\mathcal{S}$}
& \multicolumn{2}{c}{ $\mathcal{S}\rightarrow\mathcal{V}$}
& \multicolumn{2}{c}{ $\mathcal{I}\rightarrow\mathcal{S}$}
& \multicolumn{2}{c}{ $\mathcal{S}\rightarrow\mathcal{I}$}
& \multicolumn{2}{c}{ $\mathcal{V}\rightarrow\mathcal{S}$}
& \multicolumn{2}{c}{ $\mathcal{S}\rightarrow\mathcal{V}$}
& \multicolumn{2}{c}{ $\mathcal{I}\rightarrow\mathcal{S}$}
& \multicolumn{2}{c}{ $\mathcal{S}\rightarrow\mathcal{I}$} \\
\cline{1-19}

MAFE & BSSRM & SOTRA
& R@1 & AP & R@1 & AP
& R@1 & AP & R@1 & AP
& R@1 & AP & R@1 & AP
& R@1 & AP & R@1 & AP \\
\hline

& &
& 54.42 & 62.29
& 55.10 & 63.04
& 49.32 & 57.33
& 50.79 & 58.56
& 51.47 & 59.81
& 50.22 & 58.30
& 51.58 & 59.31
& 50.79 & 58.56 \\

& \checkmark & \checkmark
& 55.22 & 62.99
& 55.44 & 63.21
& 51.02 & 58.76
& 52.36 & 59.68
& 55.33 & 63.13
& 57.37 & 64.94
& 51.59 & 59.74
& 54.42 & 61.60 \\

\checkmark & \checkmark & \checkmark
& \textbf{56.12} & \textbf{63.61}
& \textbf{57.34} & \textbf{64.99}
& \textbf{53.97} & \textbf{61.12}
& \textbf{54.65} & \textbf{61.68}
& \textbf{57.48} & \textbf{64.80}
& \textbf{58.50} & \textbf{65.92}
& \textbf{53.96} & \textbf{61.32}
& \textbf{55.21} & \textbf{62.41} \\

\hline
\end{tabular}%
}

\end{table*}

\section{Experiments}
\label{experiments}



\subsection{Implementation Details}

All input images are resized to $384\times384$. During training, MASTR-Net is trained for 20 epochs with AdamW,  an initial learning rate of $1\times10^{-3}$, and a weight decay of $1\times10^{-2}$. The batch size $B$ is set to 128 for both training and testing. In SOTRA loss, the target temperature $\tau_t$, positive-pair bias $\delta$, and number of Sinkhorn iterations $K$ are set to 0.07, 2.0 and 5, respectively. The relation temperature $\tau_r$ is implicitly determined by the learnable logit scale and regulates the sharpness of the predicted relation distributions. The coefficients $\lambda_1$ and $\lambda_2$ in Eq.~(16) are set to 1. All experiments are implemented in PyTorch and conducted on Ubuntu 22.04 with four NVIDIA RTX 4090 GPUs. In the modality-balanced joint training, each geographic location is randomly assigned to the visible or infrared modality within each epoch to construct batches where visible and infrared pairs each account for \(B/2\) samples and ensure balanced supervision from the two aerial modalities.

\subsection{Comparison with State-of-the-Art Methods}

MASTR-Net is compared with state-of-the-art DVGL methods, including Sample4Geo \cite{deuser2023sample4geo}, MEAN \cite{chen2024multi}, MFRGN \cite{wang2024mfrgn}, CAMP \cite{wu2024camp}, DAC \cite{xia2024enhancing}, and SURFNet \cite{liu2026surfnet}. 
The MASTR-Net trained solely on the corresponding aerial modality and tested on the same modality and the variant trained on dual aerial modalities using the modality-balanced joint training strategy (i.e., \(\mathcal{V}\oplus\mathcal{I}\rightarrow\mathcal{S}\) and \(\mathcal{S}\rightarrow\mathcal{V}\oplus\mathcal{I}\)) but tested on the identical corresponding aerial modality are denoted as MASTR-Net\(^{\dagger}\) and MASTR-Net\(^{*}\), respectively.

\textbf{Comparison on IRVSGeo.}
For visible retrieval, MASTR-Net$^{*}$ is jointly trained on
visible--satellite and infrared--satellite image pairs using
modality-balanced sampling, and evaluated using only visible
aerial and satellite images.
As reported in Table~\ref{tab:IRVSGeo_comparison},
it achieves 57.48\% R@1 and 64.80\% AP when tested on
$\mathcal{V}\rightarrow\mathcal{S}$,
exceeding DAC in R@1 and CAMP in AP by
2.38 and 1.88 percentage points, respectively.
When tested on $\mathcal{S}\rightarrow\mathcal{V}$,
it obtains 58.50\% R@1 and 65.92\% AP, surpassing MFRGN by
3.06 and 2.65 percentage points, respectively.
Its average R@1 and AP across the two retrieval directions are
57.99\% and 65.36\%, which exceed those of DAC by
2.89 and 2.79 percentage points, respectively.
Compared with MASTR-Net$^{\dagger}$, modality-balanced joint
training improves the average visible R@1 and AP by
1.26 and 1.06 percentage points, respectively.

For infrared retrieval, MASTR-Net$^{*}$ is jointly trained on
visible--satellite and infrared--satellite image pairs using
modality-balanced sampling, and evaluated using only infrared
aerial and satellite images.
As reported in Table~\ref{tab:IRVSGeo_comparison},
it achieves 53.96\% R@1 and 61.32\% AP when tested on
$\mathcal{I}\rightarrow\mathcal{S}$, exceeding DAC by
1.58 and 1.56 percentage points, respectively.
When tested on $\mathcal{S}\rightarrow\mathcal{I}$,
it obtains 55.21\% R@1 and 62.41\% AP, surpassing SURFNet by
3.63 and 3.48 percentage points, respectively.
Its average R@1 and AP across the two retrieval directions are
54.59\% and 61.87\%, which exceed those of DAC by
2.89 and 2.73 percentage points, respectively.
Compared with MASTR-Net$^{\dagger}$, modality-balanced joint
training improves the average infrared R@1 and AP by
0.28 and 0.47 percentage points, respectively.


\textbf{Scene-wise Results.}
As shown in Tables~\ref{tab:scene_cropland}-\ref{tab:scene_urban},
the performance of MASTR-Net varies with scene characteristics but
remains competitive across all four categories. As reported in Table~\ref{tab:scene_cropland},
under single-modality training, MASTR-Net$^{\dagger}$ achieves
the highest R@1 and AP of 65.26\% and 71.22\%, respectively,
on Cropland when tested on
$\mathcal{S}\rightarrow\mathcal{V}$.
Its R@1 on $\mathcal{V}\rightarrow\mathcal{S}$ is 67.37\%,
matching MFRGN.
With modality-balanced joint training,
MASTR-Net$^{*}$ achieves 68.42\% R@1 and 73.27\% AP when tested on
$\mathcal{V}\rightarrow\mathcal{S}$,
exceeding DAC in R@1 and MFRGN in AP by
0.53 and 0.27 percentage points, respectively.
On $\mathcal{I}\rightarrow\mathcal{S}$,
it obtains 60.53\% R@1 and 65.70\% AP,
surpassing DAC by 2.64 and 1.66 percentage points, respectively.
On $\mathcal{S}\rightarrow\mathcal{I}$,
its R@1 and AP reach 58.95\% and 65.45\%,
which improve upon MASTR-Net$^{\dagger}$ by
1.06 and 0.62 percentage points, respectively.
However, its R@1 and AP on
$\mathcal{S}\rightarrow\mathcal{V}$ are 63.68\% and 70.49\%,
which are lower than those of MASTR-Net$^{\dagger}$ by
1.58 and 0.73 percentage points, respectively.

\begin{table*}[!t]
\centering

\begin{minipage}[t]{0.49\textwidth}
\centering
\caption{Sensitivity analysis of $\lambda_1$ for
MASTR-Net$^{\dagger}$ on visible and infrared modalities
of IRVSGeo.}
\label{tab:lambda1_dagger}

\resizebox{\linewidth}{!}{
\begin{tabular}{c|c|cccc}
\toprule
Modality
& $\lambda_1$
& $\mathcal{A}\rightarrow\mathcal{S}$ R@1
& $\mathcal{A}\rightarrow\mathcal{S}$ AP
& $\mathcal{S}\rightarrow\mathcal{A}$ R@1
& $\mathcal{S}\rightarrow\mathcal{A}$ AP
\\\hline

\multirow{5}{*}{Visible}
&0.4&56.23&63.51&55.78&63.53\\
&0.6&\textbf{57.02}&\textbf{64.20}&55.78&63.64\\
&0.8&56.23&63.55&54.99&62.80\\
&1.0&56.12&63.61&57.34&64.99\\
&1.2&55.56&63.19&\textbf{57.59}&\textbf{65.09}\\

\hline

\multirow{5}{*}{Infrared}
&0.4&51.92&59.51&53.17&60.39\\
&0.6&52.15&59.62&52.49&59.96\\
&0.8&52.15&59.66&53.40&60.48\\
&1.0&\textbf{53.97}&\textbf{61.12}&\textbf{54.65}&\textbf{61.68}\\
&1.2&53.29&60.47&53.40&60.45\\

\hline
\end{tabular}
}
\end{minipage}
\hfill
\begin{minipage}[t]{0.49\textwidth}
\centering
\caption{Sensitivity analysis of $\lambda_2$ for
MASTR-Net$^{\dagger}$ on visible and infrared modalities
of IRVSGeo.}
\label{tab:lambda2_dagger}

\resizebox{\linewidth}{!}{
\begin{tabular}{c|c|cccc}
\toprule
Modality
& $\lambda_2$
& $\mathcal{A}\rightarrow\mathcal{S}$ R@1
& $\mathcal{A}\rightarrow\mathcal{S}$ AP
& $\mathcal{S}\rightarrow\mathcal{A}$ R@1
& $\mathcal{S}\rightarrow\mathcal{A}$ AP
\\\hline

\multirow{5}{*}{Visible}
&0.4&\textbf{56.68}&\textbf{64.02}&57.14&64.71\\
&0.6&55.32&62.95&\textbf{58.39}&\textbf{65.63}\\
&0.8&55.32&63.02&55.21&63.15\\
&1.0&56.12&63.61&57.34&64.99\\
&1.2&55.66&63.36&56.23&63.94\\

\hline

\multirow{5}{*}{Infrared}
&0.4&50.79&58.38&53.06&60.51\\
&0.6&51.47&58.87&52.83&60.23\\
&0.8&51.58&59.19&53.28&60.29\\
&1.0&\textbf{53.97}&\textbf{61.12}&\textbf{54.65}&\textbf{61.68}\\
&1.2&51.81&59.29&53.51&60.82\\

\hline
\end{tabular}
}
\end{minipage}

\par\vspace{1.5em}

\begin{minipage}[t]{0.49\textwidth}
\centering
\caption{Sensitivity analysis of $\lambda_1$ for
MASTR-Net$^{*}$ on visible and infrared modalities
of IRVSGeo.}
\label{tab:lambda1_star}

\resizebox{\linewidth}{!}{
\begin{tabular}{c|c|cccc}
\hline
Modality
& $\lambda_1$
& $\mathcal{A}\rightarrow\mathcal{S}$ R@1
& $\mathcal{A}\rightarrow\mathcal{S}$ AP
& $\mathcal{S}\rightarrow\mathcal{A}$ R@1
& $\mathcal{S}\rightarrow\mathcal{A}$ AP
\\\hline

\multirow{5}{*}{Visible}
&0.4&55.56&63.23&\textbf{58.73}&\textbf{66.15}\\
&0.6&\textbf{58.84}&\textbf{65.98}&57.37&65.08\\
&0.8&57.26&64.54&58.61&65.98\\
&1.0&57.48&64.80&58.50&65.92\\
&1.2&56.68&64.27&57.25&64.86\\

\hline

\multirow{5}{*}{Infrared}
&0.4&53.51&60.98&54.98&62.06\\
&0.6&53.63&61.00&54.54&61.85\\
&0.8&53.97&61.44&54.31&61.75\\
&1.0&53.96&61.32&55.21&62.41\\
&1.2&\textbf{55.66}&\textbf{62.68}&\textbf{55.44}&\textbf{62.58}\\

\hline
\end{tabular}
}
\end{minipage}
\hfill
\begin{minipage}[t]{0.49\textwidth}
\centering
\caption{Sensitivity analysis of $\lambda_2$ for
MASTR-Net$^{*}$ on visible and infrared modalities
of IRVSGeo.}
\label{tab:lambda2_star}

\resizebox{\linewidth}{!}{
\begin{tabular}{c|c|cccc}
\hline
Modality
& $\lambda_2$
& $\mathcal{A}\rightarrow\mathcal{S}$ R@1
& $\mathcal{A}\rightarrow\mathcal{S}$ AP
& $\mathcal{S}\rightarrow\mathcal{A}$ R@1
& $\mathcal{S}\rightarrow\mathcal{A}$ AP
\\\hline

\multirow{5}{*}{Visible}
&0.4&57.02&64.54&57.93&65.50\\
&0.6&56.57&63.99&54.42&62.66\\
&0.8&\textbf{58.73}&\textbf{65.69}&57.48&64.93\\
&1.0&57.48&64.80&\textbf{58.50}&\textbf{65.92}\\
&1.2&56.57&64.00&55.55&63.51\\

\hline

\multirow{5}{*}{Infrared}
&0.4&54.31&61.64&54.08&61.49\\
&0.6&\textbf{55.32}&\textbf{62.56}&54.64&61.72\\
&0.8&54.08&61.44&\textbf{55.55}&\textbf{62.64}\\
&1.0&53.96&61.32&55.21&62.41\\
&1.2&54.08&61.41&54.42&61.70\\

\hline
\end{tabular}
}
\end{minipage}

\end{table*}

As reported in Table~\ref{tab:scene_coastal},
MASTR-Net$^{\dagger}$ achieves the highest visible R@1 and AP
on Coastal in both retrieval directions.
When tested on $\mathcal{V}\rightarrow\mathcal{S}$,
it obtains 48.10\% R@1 and 55.12\% AP,
exceeding SURFNet by 0.63 and 0.59 percentage points, respectively.
On $\mathcal{S}\rightarrow\mathcal{V}$,
it reaches 46.84\% R@1 and 55.51\% AP,
surpassing Sample4Geo by 2.54 and 1.85 percentage points, respectively.
Although MASTR-Net$^{*}$ has lower visible retrieval performance
than MASTR-Net$^{\dagger}$ in both directions,
its R@1 and AP on $\mathcal{S}\rightarrow\mathcal{V}$
are 45.57\% and 54.44\%, which exceed those of Sample4Geo by
1.27 and 0.78 percentage points, respectively.
For infrared retrieval, MASTR-Net$^{*}$ with modality-balanced
joint training achieves 51.26\% R@1 and 57.57\% AP on
$\mathcal{I}\rightarrow\mathcal{S}$,
surpassing MFRGN by 3.16 and 3.01 percentage points, respectively.
On $\mathcal{S}\rightarrow\mathcal{I}$,
it obtains 42.41\% R@1 and 50.78\% AP,
matching MFRGN in R@1 and exceeding DAC in AP by
0.61 percentage points.

As shown in Table~\ref{tab:scene_forest},
the two MASTR-Net variants collectively achieve the best results
in seven of the eight metrics on Forest.
When tested on $\mathcal{V}\rightarrow\mathcal{S}$,
MASTR-Net$^{\dagger}$ obtains 64.97\% R@1 and 71.14\% AP,
exceeding MEAN by 2.79 and 2.37 percentage points, respectively.
With modality-balanced joint training,
MASTR-Net$^{*}$ achieves the same R@1 of 64.97\%
and an AP of 71.06\% in this direction.
On $\mathcal{S}\rightarrow\mathcal{V}$,
MASTR-Net$^{*}$ reaches 65.20\% R@1 and 71.34\% AP,
surpassing DAC by 2.32 and 1.92 percentage points, respectively.
On $\mathcal{I}\rightarrow\mathcal{S}$,
it achieves 62.88\% R@1 and 68.38\% AP,
exceeding DAC by 4.64 and 3.87 percentage points, respectively.
On $\mathcal{S}\rightarrow\mathcal{I}$,
it obtains 62.18\% R@1 and 68.65\% AP,
improving upon MASTR-Net$^{\dagger}$ by
1.86 and 1.75 percentage points, respectively.
Compared with DAC in this direction,
its AP is 0.08 percentage points higher,
while its R@1 is 0.23 percentage points lower.

As reported in Table~\ref{tab:scene_urban},
MASTR-Net$^{\dagger}$ achieves the highest visible R@1 and AP
on Urban when tested on $\mathcal{V}\rightarrow\mathcal{S}$,
with 69.90\% and 74.27\%, respectively.
These results exceed MEAN in R@1 and Sample4Geo in AP by
2.91 and 2.51 percentage points, respectively.
With modality-balanced joint training,
MASTR-Net$^{*}$ obtains 66.99\% R@1 and 71.76\% AP on
$\mathcal{V}\rightarrow\mathcal{S}$,
and 56.31\% R@1 and 63.62\% AP on
$\mathcal{S}\rightarrow\mathcal{V}$.
Both metrics are lower than those of MASTR-Net$^{\dagger}$
in the two visible retrieval directions.
For infrared retrieval,
MASTR-Net$^{*}$ achieves 63.11\% R@1 and 67.14\% AP on
$\mathcal{I}\rightarrow\mathcal{S}$,
exceeding DAC by 1.94 and 1.54 percentage points, respectively.
On $\mathcal{S}\rightarrow\mathcal{I}$,
it obtains 60.19\% R@1 and 66.41\% AP,
surpassing DAC by 3.88 and 3.26 percentage points, respectively.
Compared with MASTR-Net$^{\dagger}$,
its infrared R@1 and AP improve by
3.89 and 4.02 percentage points, respectively, on
$\mathcal{I}\rightarrow\mathcal{S}$,
and by 4.85 and 4.45 percentage points, respectively, on
$\mathcal{S}\rightarrow\mathcal{I}$.

\subsection{Ablation Study}

To evaluate the contributions of the proposed components,
BSSRM and SOTRA loss are first added together to the encoder,
followed by MAFE.
Table~\ref{tab:component_ablation} reports the results under
single-modality training for MASTR-Net$^{\dagger}$ and
modality-balanced joint training on visible--satellite and
infrared--satellite image pairs for MASTR-Net$^{*}$.
Compared with the encoder, adding BSSRM and SOTRA loss
under single-modality training improves visible R@1 by
0.80 and 0.34 percentage points when tested on
$\mathcal{V}\rightarrow\mathcal{S}$ and
$\mathcal{S}\rightarrow\mathcal{V}$, respectively.
The corresponding infrared R@1 gains are
1.70 and 1.57 percentage points on
$\mathcal{I}\rightarrow\mathcal{S}$ and
$\mathcal{S}\rightarrow\mathcal{I}$, respectively.
Under modality-balanced joint training,
the R@1 gains over the encoder are
3.86 and 7.15 percentage points when tested on
$\mathcal{V}\rightarrow\mathcal{S}$ and
$\mathcal{S}\rightarrow\mathcal{V}$, respectively.
On $\mathcal{S}\rightarrow\mathcal{I}$,
R@1 and AP improve by 3.63 and 3.04 percentage points,
respectively.
On $\mathcal{I}\rightarrow\mathcal{S}$,
R@1 remains nearly unchanged at 51.59\% compared with 51.58\%,
while AP increases from 59.31\% to 59.74\%.

Adding MAFE to the configuration with BSSRM and SOTRA loss
improves both R@1 and AP in all four retrieval directions
under both training settings.
Under single-modality training,
MASTR-Net$^{\dagger}$ achieves additional visible R@1 gains of
0.90 and 1.90 percentage points when tested on
$\mathcal{V}\rightarrow\mathcal{S}$ and
$\mathcal{S}\rightarrow\mathcal{V}$, respectively.
The corresponding infrared R@1 gains are
2.95 and 2.29 percentage points on
$\mathcal{I}\rightarrow\mathcal{S}$ and
$\mathcal{S}\rightarrow\mathcal{I}$, respectively.
Under modality-balanced joint training,
MASTR-Net$^{*}$ achieves additional visible R@1 gains of
2.15 and 1.13 percentage points when tested on
$\mathcal{V}\rightarrow\mathcal{S}$ and
$\mathcal{S}\rightarrow\mathcal{V}$, respectively.
Its infrared R@1 gains are
2.37 and 0.79 percentage points on
$\mathcal{I}\rightarrow\mathcal{S}$ and
$\mathcal{S}\rightarrow\mathcal{I}$, respectively.
Compared with the encoder under the corresponding training setting,
the complete model achieves maximum R@1 gains of
2.24 and 4.65 percentage points for visible and infrared retrieval,
respectively, under single-modality training.
Under modality-balanced joint training,
the corresponding maximum gains are
8.28 and 4.42 percentage points, respectively.

\begin{table*}[t]
\centering
\caption{Comparison with state-of-the-art methods on IR-VL328 and
CVGL-RGBT under the individually trained protocol. Avg. denotes the
average over the two retrieval directions.}
\label{tab:infrared_comparison}
\setlength{\tabcolsep}{3.0pt}
\renewcommand{\arraystretch}{1.08}
\begin{tabular}{l c cccccc cccccc}
\hline
\multirow{3}{*}{Model}
& \multirow{3}{*}{Venue}
& \multicolumn{6}{c}{IR-VL328}
& \multicolumn{6}{c}{CVGL-RGBT} \\
\cline{3-14}
&
& \multicolumn{2}{c}{$\mathcal{A}\rightarrow\mathcal{S}$}
& \multicolumn{2}{c}{$\mathcal{S}\rightarrow\mathcal{A}$}
& \multicolumn{2}{c}{Avg.}
& \multicolumn{2}{c}{$\mathcal{A}\rightarrow\mathcal{S}$}
& \multicolumn{2}{c}{$\mathcal{S}\rightarrow\mathcal{A}$}
& \multicolumn{2}{c}{Avg.} \\
\cline{3-14}
&
& R@1 & AP & R@1 & AP & R@1 & AP
& R@1 & AP & R@1 & AP & R@1 & AP \\
\hline
Sample4Geo\cite{deuser2023sample4geo} & ICCV'2023
& 68.19 & 73.43 & 75.78 & 63.92 & 71.99 & 68.68
& 89.33 & 91.46 & 88.00 & 90.65 & 88.67 & 91.06 \\

MEAN\cite{chen2024multi} & TGRS'2025
& 63.75 & 70.11 & 68.75 & 57.36 & 66.25 & 63.74
& 81.33 & 85.31 & 86.66 & 89.45 & 84.00 & 87.38 \\

MFRGN\cite{wang2024mfrgn} & ACM MM'2024
& 67.81 & 73.25 & 78.12 & 59.33 & 72.97 & 66.29
& 88.00 & 90.50 & 84.00 & 87.48 & 86.00 & 88.99 \\

CAMP\cite{wu2024camp} & TGRS'2024
& 67.04 & 72.60 & 71.87 & 63.37 & 69.46 & 67.99
& 85.33 & 88.03 & 83.33 & 86.42 & 84.33 & 87.23 \\

DAC\cite{xia2024enhancing} & TCSVT'2024
& 71.15 & 75.56 & 75.00 & 65.38 & 73.08 & 70.47
& \textbf{91.33} & \textbf{93.12}
& \textbf{93.33} & 94.71
& \textbf{92.33} & \textbf{93.92} \\

SURFNet\cite{liu2026surfnet} & TGRS'2026
& 65.19 & 70.87 & 71.88 & 59.01 & 68.54 & 64.94
& 88.00 & 90.56 & 89.33 & 91.13 & 88.67 & 90.85 \\
\hline
MASTR-Net & -
& \textbf{74.09} & \textbf{78.69}
& \textbf{79.68} & \textbf{71.58}
& \textbf{76.89} & \textbf{75.14}
& 90.00 & 92.13
& \textbf{93.33} & \textbf{94.80}
& 91.67 & 93.47 \\
\hline
\end{tabular}%
\end{table*}

\subsection{Further Analysis}

\textbf{Sensitivity Analysis of $\lambda_{1}$.}
As shown in Tables~\ref{tab:lambda1_dagger} and~\ref{tab:lambda1_star}, the sensitivity of MASTR-Net to the loss balancing coefficient $\lambda_{1}$ is investigated under visible and infrared modalities. The coefficient is
varied from $0.4$ to $1.2$, while other settings remain unchanged.
For both modalities, MASTR-Net$^{*}$ consistently achieves better
performance than MASTR-Net$^{\dagger}$ under most coefficient settings,
which demonstrates the effectiveness of the modality-balanced optimization
strategy. The improvements are consistently observed in the retrieval
directions $\mathcal{A}\rightarrow\mathcal{S}$ and
$\mathcal{S}\rightarrow\mathcal{A}$, which indicates that the proposed
strategy benefits cross-view matching from different retrieval
perspectives.

Meanwhile, both variants exhibit relatively stable performance with
respect to the variation of $\lambda_{1}$, which shows the robustness of
the proposed framework against loss weight changes. Although individual
metrics may achieve their best values at different coefficients,
excessively small or large $\lambda_{1}$ values cannot simultaneously
maintain balanced retrieval performance across different modalities and
retrieval directions. Therefore, $\lambda_{1}=1.0$ is adopted in subsequent
experiments based on the overall performance consistency and
cross-modality balance.

\textbf{Sensitivity Analysis of $\lambda_{2}$.}
Table~\ref{tab:lambda2_dagger} and Table~\ref{tab:lambda2_star}
present the sensitivity analysis of MASTR-Net$^{\dagger}$ and
MASTR-Net$^{*}$ with respect to $\lambda_{2}$ under visible and
infrared modalities. The results show that the performance of both
variants remains relatively stable when $\lambda_{2}$ is varied,
which indicates that the proposed framework is robust to moderate changes
of the loss weighting.

Different modalities and retrieval directions exhibit different preferred
coefficient values. For instance, MASTR-Net$^{*}$ achieves its highest
visible performance when tested on
$\mathcal{A}\rightarrow\mathcal{S}$ at $\lambda_{2}=0.8$, while the
best performance when tested on
$\mathcal{S}\rightarrow\mathcal{A}$ is obtained at $\lambda_{2}=1.0$.
Similarly, the optimal settings in the infrared modality vary across
retrieval directions. Although certain coefficients may improve individual
metrics, they do not always provide balanced gains across all modalities
and retrieval directions.

Therefore, $\lambda_{2}=1.0$ is adopted in subsequent experiments as a
reasonable trade-off between different retrieval directions and modalities.
The comparison between MASTR-Net$^{\dagger}$ and MASTR-Net$^{*}$ under
different settings further verifies the effectiveness of the proposed
optimization strategy.

\textbf{Evaluation on Additional Infrared Benchmarks.}
To further evaluate the applicability of MASTR-Net on additional
infrared benchmarks, experiments are conducted on IR-VL328 and
CVGL-RGBT. As reported in
Table~\ref{tab:infrared_comparison}, MASTR-Net achieves the best
performance across all metrics on IR-VL328, with average R@1 and AP
of 76.89\% and 75.14\%, respectively. These results exceed those of
DAC by 3.81 and 4.67 percentage points, respectively. Notably,
its AP on $\mathcal{S}\rightarrow\mathcal{A}$
reaches 71.58\%, which outperforms DAC by 6.20 percentage points.

On CVGL-RGBT, MASTR-Net achieves 93.33\% R@1 and 94.80\% AP when
tested on $\mathcal{S}\rightarrow\mathcal{A}$, which matches
the best R@1 result and improves the best AP result by 0.09 percentage points. Its
average performance, which reaches 91.67\% R@1 and 93.47\% AP, remains
comparable to that of DAC. These results demonstrate that MASTR-Net
maintains strong performance across benchmarks with different scene
distributions and imaging conditions, which indicates that its
effectiveness is not limited to IRVSGeo.

\section{Conclusion}
\label{conclusions}
This work addresses the practical day-night drone-view geo-localization problem under varying illumination conditions. To overcome the limitations of existing benchmarks, IRVSGeo, a geographically aligned benchmark that contains visible aerial images, infrared aerial images, and corresponding satellite reference images from 8,820 locations, is constructed. IRVSGeo enables a unified performance evaluation in the day and night at the same geographic location. Furthermore, MASTR-Net is proposed, in which MAFE, the BSSRM module, and SOTRA loss are integrated to address modality discrepancies and cross-view structural discrepancies. Extensive experiments on IRVSGeo and two additional infrared benchmarks demonstrate the effectiveness and generalizability of the proposed MASTR-Net. The proposed benchmark and MASTR-Net provide a unified foundation for the advancement of reliable DVGL systems under varying illumination and sensing conditions. Future work will explore more challenging DVGL scenarios, for which IRVSGeo will be extended with multi-temporal observations, larger geographic coverage, and additional sensing modalities.

\begingroup
\footnotesize
\bibliographystyle{IEEEtran}
\bibliography{reference}
\endgroup

\end{document}